\documentclass[lettersize,journal]{IEEEtran}
\usepackage{amsmath,amsfonts}
\usepackage{array}
\usepackage{textcomp}
\usepackage{stfloats}
\usepackage{url}
\usepackage{verbatim}
\usepackage{graphicx}
\usepackage{cite}
\usepackage{cuted}
\usepackage{hyperref}
\usepackage{amsthm}
\usepackage{subcaption}
\usepackage{soul}
\usepackage{xcolor}
\usepackage[linesnumbered,ruled,vlined]{algorithm2e}

\usepackage[switch]{lineno}
\usepackage{multirow}
\usepackage{balance}
\usepackage{booktabs}

\begin{document}




\title{A Transfer Learning-Driven Methodology for Efficient and Sustainable Indoor Localization Using VLC}

\author{Masood Jan, Wafa Njima, Xun Zhang, Alexander Artemenko

\thanks{This paper extends our earlier work \cite{jan2025transfer}, which was presented at the IEEE VTC2025-Spring Conference in Oslo, Norway, in June 2025. In the previous work, we introduced a transfer learning-based approach for VLC-driven indoor localization to address environmental variability and its effect on accuracy. The current extended version refines the methodology and introduces a unified evaluation metric 
enabling a holistic assessment across energy efficiency, adaptability, data usage, and accuracy. }

\thanks{Masood Jan is the corresponding author. He is with the Doctoral School of Computer Science, Telecommunications and Electronics (EDITE), Sorbonne University, Paris, France and affiliated with the LISITE Laboratory, Institut Supérieur d'Électronique de Paris, Paris, France (e-mail: \mbox{masoodjann@gmail.com}).}
\thanks{ Wafa Njima and Xun Zhang are with the LISITE Laboratory, Institut Supérieur d'Électronique de Paris, Paris, France. (e-mail: {wafa.njima, xun.zhang}@isep.fr). Xun Zhang is also affiliated with the Université Paris-Saclay, LISV Laboratory, France.}
\thanks{ Alexander Artemenko is with the sector research of Robert Bosch GmbH in Stuttgart, Germany.  (e-mail: alexander.artemenko@de.bosch.com).}
\thanks{\textcopyright{} 2025 IEEE. Personal use of this material is permitted. Permission from IEEE must be obtained for all other uses, in any current or future media, including reprinting/republishing this material for advertising or promotional purposes, creating new collective works, for resale or redistribution to servers or lists, or reuse of any copyrighted component of this work in other works.}

\thanks{Published in IEEE Internet of Things Journal, vol. 13, no. 2, pp. 3106--3122, 2026. DOI: \url{https://doi.org/10.1109/JIOT.2025.3633511}.}

}

\maketitle

\begin{abstract}
The rapid expansion of the Internet of Things enables seamless collaboration among connected devices, making indoor localization a critical component of industrial automation. However, conventional localization techniques struggle in dynamic environments. These methods depend on extensive, frequent data collection and environment-specific calibration, limiting scalability, interoperability, and effective use of prior research while imposing significant energy penalties that hinder practical deployment. 
To address these challenges, we propose an energy-efficient transfer learning (TL) based approach for visible light communication based indoor localization. Our method leverages TL to tackle environmental variability, including lighting fluctuations and physical obstacles that typically degrade localization performance,  while simultaneously reducing computational overhead and energy consumption compared to conventional approaches. We also introduce a novel model efficiency (ME) metric, designed to integrate
localization accuracy, energy efficiency, data efficiency, and transfer gain into a single evaluative measure for comprehensive optimization.
Evaluations on a real-world dataset collected from a BOSCH factory 
demonstrate that our model achieves a 47\% improvement in localization accuracy compared to a conventional model that does not utilize TL, and achieves up to 67.5\% ME with TL models, compared to only 10.4\% in conventional models under high-noise conditions.
These results underscore the potential of our approach to deliver a highly efficient and scalable indoor localization system suitable for energy-constrained industrial applications.

\end{abstract}

\section{Introduction}
\IEEEPARstart{T}{he} convergence of Industry 4.0 technologies and ubiquitous connectivity solutions has elevated indoor localization from a supportive tool to a critical infrastructure component in modern industrial ecosystems \cite{aziz2024enhancing}.
As manufacturing and logistics environments grow increasingly complex with autonomous mobile robots \cite{knitt2023benchmarking}, 
digital twin implementations \cite{ullah2025digital}, and artificial intelligence (AI)-driven process automation \cite{marasovic2024indoor}, the need for precise, reliable and 
efficient indoor localization system has become critical. This rising demand is driving innovations in location-based services (LBS), transforming industrial operations through enhanced asset tracking, real-time monitoring, predictive maintenance, and streamlined workflow automation \cite{pan2008transfer}.

Despite significant technological advancements, current indoor localization systems face critical performance challenges. These include accuracy, energy consumption, computational complexity, data efficiency, and adaptability to changing environments. These factors have become the primary bottleneck, fundamentally constraining scalability and industrial adoption. For example, energy-intensive algorithms are impractical for battery-powered sensor nodes, and high-speed production lines cannot accommodate lengthy data collection or retraining phases \cite{zheng2016energy}—even if such systems offer high accuracy \cite{ahmed2024unified, chen2022few}.
Wi-Fi-based systems offer relatively low power consumption 
\cite{dai2023survey}, but their performance is highly susceptible to environmental interference, resulting in degraded accuracy under dynamic conditions \cite{gonzalez2019energy}. Bluetooth Low Energy (BLE) improves energy efficiency further by consuming as little as 10.1 µA in sleep mode and extending device runtime significantly compared to Wi-Fi \cite{lindemann2016indoor}—but suffers from limited range and poor signal stability in changing environments \cite{milano2024ble, wang2025ble}. Both Wi-Fi and BLE require dense deployment of access points or battery-powered beacons, leading to high installation and maintenance costs \cite{xiao2016survey}. Ultra-Wideband (UWB) technology, while delivering high accuracy, faces energy challenges 
\cite{khriji2022design}, high device costs, and limited adaptability to environmental changes \cite{schmidt2021detection}. Radio frequency (RF) systems are further hindered by electromagnetic interference (EMI) from industrial machinery, which can cause signal attenuation exceeding 15 dBm and necessitate increased transmission power to maintain connectivity \cite{aziz2025comprehensive}.


These challenges highlight the need for localization solutions that deliver both high accuracy and operational efficiency, while leveraging cost-effective infrastructure. Visible Light Communication (VLC) has emerged as a promising candidate, utilizing existing Light-Emitting Diode (LED) lighting for both communication and localization. VLC offers centimeter-level precision and can reduce energy consumption by up to 60\% compared to dedicated RF systems \cite{almadani2020visible, shao2016joint}. The technology is inherently immune to EMI, eliminating the energy penalties associated with increased transmission power in industrial environments \cite{miranda2023review, wang2024application}, and its short-wavelength, line-of-sight (LOS) nature keeps computational demands low \cite{ma2023centimeter}.
Nevertheless, VLC systems face their own environmental adaptability challenges, distinct from those of RF-based solutions. 
While RF systems struggle with unpredictable EMI and multipath fading, VLC contends with more deterministic optical issues: ambient light interference, LOS dependencies, and varying lighting conditions
\cite{cuailean2024driving, apolo2025experimental, zhao2024ultra}. However, these optical challenges are generally more predictable and manageable than the chaotic interference often encountered by RF systems in industrial settings \cite{oluwole2024exploration, santhanalakshmivlc}.
To address these challenges and enhance VLC-based localization, machine learning (ML) models have been increasingly adopted to capture complex signal patterns \cite{ghonimindoor2024}. Yet, many existing ML models rely on assumptions that limit their real-world applicability: they often require large, labeled datasets for offline training and assume the resulting model will remain effective across different times and deployment contexts \cite{pan2008transfer}. In dynamic industrial environments, where lighting and layouts frequently change, these assumptions break down, leading to rapid degradation in model performance.
Several adaptive ML strategies have been proposed to overcome these limitations. Federated Learning (FL) enables decentralized model training across multiple devices, preserving data privacy and capturing diverse conditions \cite{jan2024privacy}, but requires frequent updates in highly dynamic settings. Continuous Learning (CL) supports incremental adaptation over time \cite{wickramasinghe2023continual, zhao2025high}, but is vulnerable to catastrophic forgetting and depends on costly, labor-intensive labeled data collection. Transfer Learning (TL) offers rapid adaptation to new conditions without extensive retraining or system downtime \cite{su2023four, zhu2024survey}; however, in existing literature TL has been widely applied in RF-based localization, and has seen limited adaptation to VLC systems, with domain-specific noise and environmental variability still insufficiently addressed.

VLC stands out as a strong candidate, but realizing its full potential will require innovative ML approaches that address both the unique optical challenges of VLC and the stringent operational constraints of Industry 4.0. This realization motivates our work to design an enhanced TL based efficient model framework for indoor based VLC localization that is simultaneously precise and adaptive. This is an extension of our previous work detailed in \cite{jan2025transfer}.
Our key contributions are as follows:


\begin{itemize}
\item\textbf{TL Framework with Dynamic Adaptability and Cross-Environment Generalization:}
We develop a novel TL framework using on a Deep Neural Network (DNN) that not only adapts effectively to dynamic indoor environments, but also requires significantly less training data—up to 30\% of what traditional models demand and assess the practical applicability of the model by evaluating its performance under varying environmental conditions without requiring additional training. 

\item\textbf{Unified Model Efficiency Metric:} 
To navigate the trade-offs between accuracy, energy use, data needs, and transfer gains, we introduce a single composite metric called \textit{Model Efficiency (ME)}. This metric guides model selection to meet diverse industrial requirements.

\item \textbf{Model Evaluation based on a real-world VLC dataset:} We use a comprehensive dataset collected from an active production line at the BOSCH manufacturing facility in Blaichach, Germany, one of Europe's leading automotive component manufacturing sites processing over 60,000 sensor readings per hour \cite{bosch2025nexeed}. This dataset captures operational complexities and signal variations, offering a robust benchmark for evaluating VLC-based indoor localization systems.

\end{itemize}

\noindent The remainder of this paper is organized as follows: Section \ref{sec:relworks} presents the related works while Section \ref{sec:sysm} depicts the problem formulation and the system model description. Section~\ref{sec:algorithm} details the proposed algorithm development. In Section \ref{sec:setup} we discuss the experimental setup and data collection process. In Section \ref{sec:results} we go through the performance evaluation and analysis before concluding in Section \ref{sec:conclusion}.

\section{Related works}
\label{sec:relworks}
Over the last few years, several studies have been published to explore various machine learning paradigms, including Federated Learning (FL), Continuous Learning (CL), and Transfer Learning (TL), in order to improve indoor localization systems. 
These efforts aim to improve positioning accuracy, reduce energy and deployment costs, and support adaptability in dynamic environments. This section reviews key contributions in terms of model adaptability, energy efficiency, and transferability, and identifies remaining gaps in applying these approaches to VLC-based systems.

\subsection{Federated Learning for Indoor Localization}


FL has gained recognition as a promising approach for collaborative model training across distributed clients. It ensures data privacy through decentralized processing and improves energy efficiency by reducing data transmission requirements \cite{nguyen2021federated}. Several studies have explored the application of FL to indoor localization tasks, enabling collaborative model training across multiple devices to capture diverse environmental conditions without centralizing data \cite{jan2024privacy, kumar2023confidentiality, etiabi2023federated}. By keeping data on each device, FL limits exposure to privacy risks. Processing and training occur locally, with only essential updates (e.g., gradients or parameters) sent to a central server, thus avoiding raw data transmission, enhancing privacy, and reducing bandwidth usage and minimizing the energy overhead typically associated with centralized data processing \cite{jan2024privacy}. FL specifically addresses adaptability challenges in indoor localization by leveraging diverse environmental conditions captured across multiple devices without centralizing sensitive data \cite{etiabi2024femloc}. FeMLoc, a federated meta-learning framework, demonstrates this adaptability by operating in two stages: collaborative meta-training on diverse datasets and rapid adaptation to new environments with minimal fine-tuning. This approach achieves up to 80.95\% improvement in localization accuracy compared to conventional neural networks while reducing training iterations by 82.21\%.
In \cite{kumar2023confidentiality}, a federated architecture for Wi-Fi fingerprinting preserves data privacy across edge devices while simultaneously improving localization accuracy and reducing the energy burden on individual devices through distributed processing. Similarly, \cite{etiabi2023federated} aggregates and shuffles building data before partitioning it into clients, enabling better model tuning, improved accuracy, and reduced communication overhead compared to traditional machine learning methods. Other works \cite{ibnatta2022indoor, 9761235, guo2023fedpos, yaya-sensors} also integrate FL as a key component in localization system design, demonstrating its potential for energy-efficient collaborative learning.
However, despite these advantages, many FL implementations tend to produce models that are closely aligned with the specific characteristics of the participating clients, which can limit their generalizability or performance when deployed on new devices operating under unseen conditions \cite{rev1p2}.
This inflexibility is problematic in dynamic settings with evolving spatial layouts or fluctuating signals, where maintaining accuracy requires costly retraining that can negate the initial energy savings. Such limitations highlight performance inequities across heterogeneous deployments and underscore critical needs for architectural innovations including transfer learning (TL) mechanisms that can maintain both accuracy and energy efficiency.

\subsection{Continuous Learning for Indoor Localization}
CL presents a solution by allowing models to incrementally update their knowledge as new data becomes available. Unlike conventional retraining, which requires full access to historical data and significant computational effort, CL enables continual adaptation over time, making it suitable for scenarios with evolving signal conditions and changing spatial layouts \cite{wickramasinghe2023continual}.
Recent surveys, such as \cite{van2022three}, provide empirical evaluations and categorization frameworks for CL strategies. A core challenge identified is \textit{catastrophic forgetting}, the tendency of models to overwrite previously learned information when learning new tasks. Mitigating this phenomenon often requires additional memory, complex update strategies, and increased computation, which may undermine CL's advantages in resource-constrained settings.
Various CL strategies have been proposed to address this issue. For instance, \cite{de2021continual} classified task-incremental approaches into three categories: replay-based (which store and replay past samples), regularization-based (which constrain weight updates), and parameter isolation (which allocate different subsets of model parameters per task). While effective to varying degrees, each approach introduces trade-offs: replay methods increase memory and latency, regularization demands frequent updates, and isolation strategies may lead to model scalability issues. Similarly, \cite{masana2022class} examined class-incremental learning techniques in image classification, showing that approaches like rehearsal and bias correction further add to computational and design complexity.
In practice, CL effectiveness depends on algorithmic design, access to large, labeled datasets, and sustained computational resources. These requirements are often scarce in industrial environments. Furthermore, the need for frequent model updates in response to new data can lead to inconsistent energy usage, making CL difficult to integrate into energy-constrained systems such as battery-powered nodes. These constraints limit the practicality of CL for industrial indoor localization, where models must be both adaptable and resource-efficient, and where access to labeled data is often limited. In this context, TL offers a more scalable and lightweight alternative, enabling efficient generalization to new environments with minimal fine-tuning, reduced energy consumption, and lower data requirements.

Although FL and CL address important aspects of model adaptability, their practical suitability for VLC-based localization differs from that of TL. FL excels at privacy-preserving, collaborative training across distributed clients, but it relies on frequent parameter exchanges and stable uplink connectivity \cite{naser2022toward, sheikholeslami2022communication}; in VLC systems the uplink channel is often constrained (e.g., IR or RF fallback, limited uplink bandwidth), and optical channel characteristics (directionality, strong ambient-light variation, non-stationary SNR) can exacerbate client heterogeneity and slow convergence \cite{yan2023rethinking}. CL supports incremental updates and can adapt to evolving spatial layouts, yet common mitigation strategies for catastrophic forgetting—replay buffers, parameter isolation, or regularization—introduce memory, latency, or model-complexity overheads that are unattractive for resource-limited VLC nodes and for deployments with limited data \cite{goodfellow2013empirical, kirkpatrick2017overcoming}.


\subsection{Transfer Learning for Indoor Localization}

The limitations of Federated and Continuous Learning particularly in terms of adaptability, data dependency, and energy efficiency, have highlighted the growing relevance of TL as a scalable and practical solution for indoor localization. TL enables the reuse of knowledge from previously trained models, substantially reducing the need for extensive retraining and large labeled datasets in new deployment contexts. This results in lower computational and energy costs, faster deployment, and improved model generalization across environments \cite{iman2023review, ali2023transfer}.
According to \cite{ying2018transfer, pan2009survey}, three critical questions underpin TL research: when to transfer, how to transfer, and what to transfer. These considerations are particularly important in resource-constrained industrial environments where the cost of adaptation must be balanced against expected performance gains.
TL techniques have been successfully implemented in RF-based localization systems such as Wi-Fi, BLE, and UWB. For instance, \cite{pan2008transfer} reused convolutional layers in Wi-Fi fingerprinting models to achieve strong accuracy with only 55\% of training data. \cite{suwannaphong2024transfer} improved BLE-based room classification by leveraging TL across multiple home environments, increasing macro F1 scores in hard-to-predict areas. Similarly, \cite{adamkiewicz2021improving} applied TL in UWB systems to reduce the need for dense fingerprinting, achieving sub-30 cm accuracy while cutting data collection efforts by 93\%.
Despite these successes, TL remains underutilized in VLC-based indoor localization, highlighting an opportunity for further investigation in this domain.
Given VLC's inherent advantages such as high bandwidth, EMI immunity, and energy efficiency via infrastructure reuse, it stands to benefit significantly from TL-enabled generalization across diverse lighting and spatial conditions. Yet, most existing VLC studies focus on deterministic modeling or hardware-based noise mitigation rather than learning-based domain adaptation. For example, \cite{zhuang2019visible} employed adaptive least squares and Kalman filtering for noise handling under fixed conditions, while \cite{zhang2023recent} used singular value decomposition (SVD) for high-precision 3D positioning, prioritizing efficiency but lacking adaptability. \cite{rekkas2023artificial} surveyed ML-based approaches for VLC, noting that while fingerprinting methods are widely studied, domain-transferability and noise robustness remain open challenges. Moreover, recent TL studies in RF-based systems rarely address VLC-specific issues such as ambient light variation, photodiode noise, or LED flicker. \cite{yang2024indoor}, for example, used domain-adversarial learning to reduce Wi-Fi localization error but did not consider optical-domain interference. Similarly, \cite{ahmed2024unified} improved LoRa/Wi-Fi models via latent space alignment, without addressing VLC-specific degradation modes.
Our work bridges this gap through an integrated contribution, spanning a noise-adaptive model architecture, a unified ME metric, and industrial scale validation, that advances the development of scalable, adaptable, and energy-efficient TL-driven VLC-based indoor localization system capable of generalizing across diverse environments.


\section{System model and problem formulation}
\label{sec:sysm}

\begin{figure*}[!t]
  \centering
  \resizebox{0.9\linewidth}{!}{\includegraphics{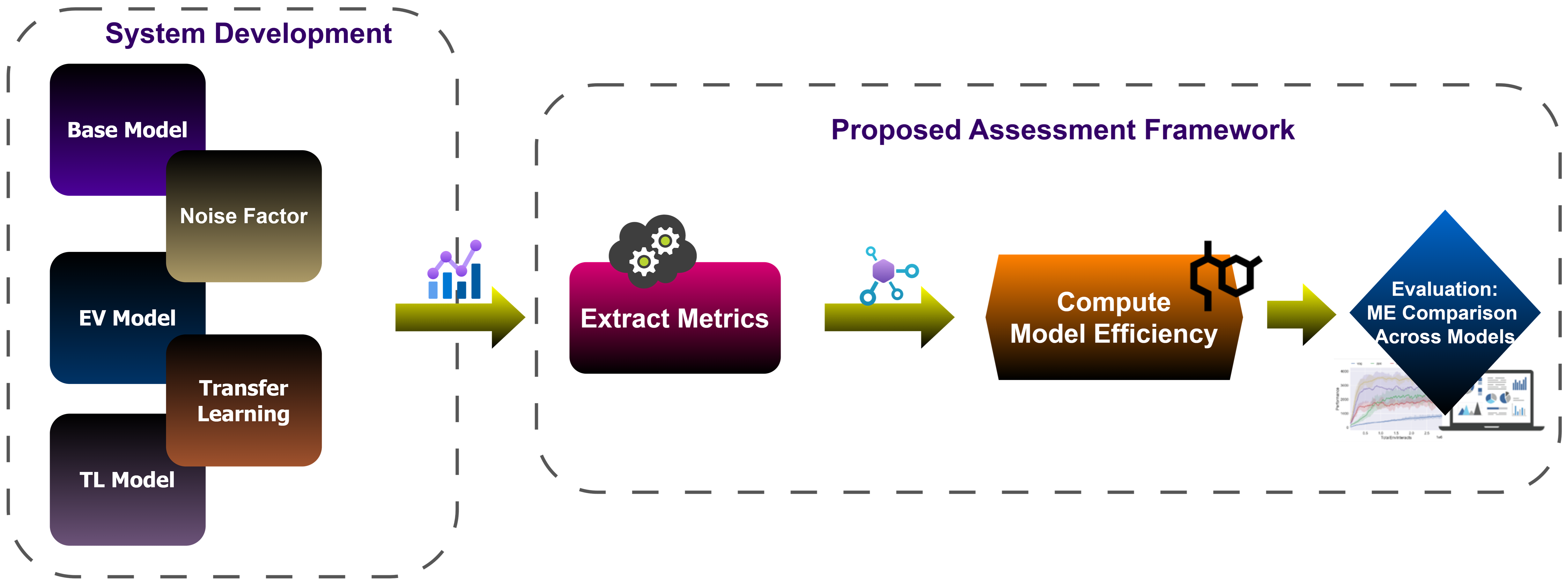}}
  \caption{System model illustrating the TL-based framework for model development and evaluation.}
  \label{fig:ME framework}
\end{figure*}

This section is organized in two parts.
First, we establish a structured approach for modeling and evaluating three DNN configurations, the Base Model, the Environmental Variation (EV) Model, and the TL Model, under varying environmental conditions. 
Although these configurations provide a foundation for understanding model adaptation, a comprehensive assessment is necessary to capture real-world deployment trade-offs. To address this, we introduce the Model Efficiency (ME) metric in \ref{ME}. It offers a unified framework to compare models across critical dimensions such as accuracy, energy consumption, data utilization, and the benefits of TL. This framework, illustrated in Fig.~\ref{fig:ME framework}, supports robust and sustainable comparison of model configurations (e.g., Base, EV, TL) under varying environmental conditions to guide deployment decisions.


\subsection{Modeling Environmental Variability and TL Framework}
\label{subsec:3A}

\begin{figure*}[!t]
  \centering
  \resizebox{0.9\linewidth}{!}{\includegraphics{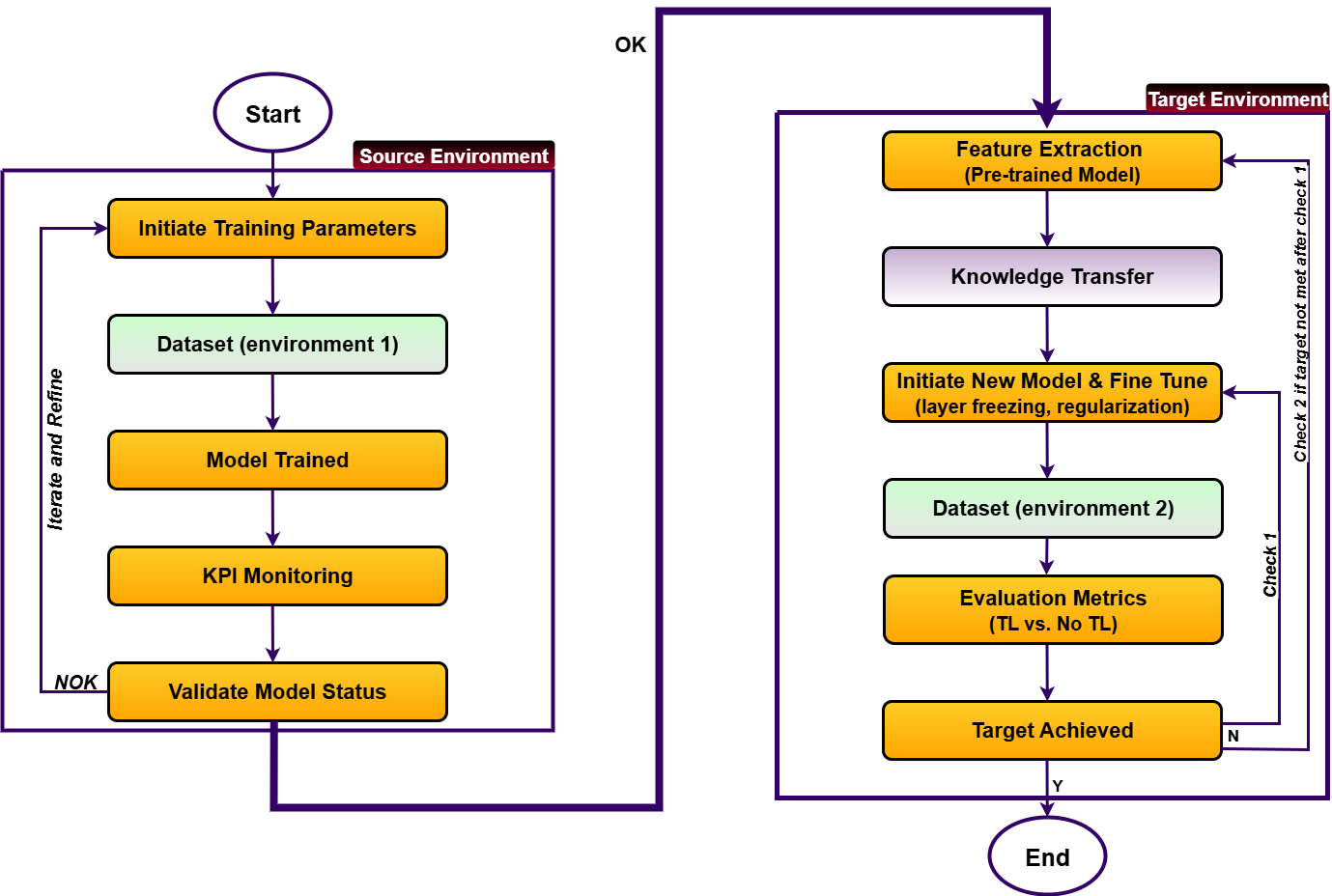}}
  \caption{Proposed TL-based localization process.}
  \label{fig:TL framework}
\end{figure*}

We implement a TL approach to predict users' coordinates (longitude and latitude) based on collected VLC data. 
As shown in Fig.~\ref{fig:TL framework}, the framework transfers knowledge from a pre-trained source model \( f_s \) to a target model \( f_t \), enabling better generalization under new environmental conditions. Our evaluation encompasses three model configurations:

\begin{enumerate}
    \item \textbf{Base Model:} The Base Model refers to the initial source model \( f_s \), trained on the raw VLC dataset. It served as the benchmark for our experiments and provided a foundation for comparison with other configurations.
    
    \item \textbf{EV Model:} The EV Model extends the Base Model by introducing a Noise Factor (NF) in the original dataset, to simulate lighting variability. This addition reflects real-world uncertainties in signal strength, addressing challenges caused by fluctuating environmental conditions.
    
    \item \textbf{TL Model:} The TL Model represents the target model \( f_t \). It initializes with the pre-trained weights of the Base Model (i.e., source model \( f_s \)) and is fine-tuned using a subset of data from the new environment. This process facilitates adaptation to environmental changes while leveraging previously learned representations.
    
\end{enumerate}

To formalize this process and underlying the domain relationships in our framework, we define the source and target domains, loss functions, and fine-tuning methodology.

\subsubsection{Problem Formulation}
Let \( D_s = \{(x_i, y_i)\}_{i=1}^{n_s} \) be the source domain with \( n_s \) samples, where each input feature vector \( x_i \in \mathbb{R}^d \) corresponds to an output label \( y_i \in \mathbb{R}^2 \), representing the longitude and latitude. Here, \( \mathbb{R}^d \) is the \( d \)-dimensional feature space (signal intensity values), and \( \mathbb{R}^2 \) is the output label space. The source task \( T_s \) is to learn a mapping function \( f_s: \mathbb{R}^d \rightarrow \mathbb{R}^2 \) from the source data. Similarly, let \( D_t = \{(x_j, y_j)\}_{j=1}^{n_t} \) be the target domain with \( n_t \) samples, where each input vector \( x_j \in \mathbb{R}^d \) corresponds to an output label \( y_j \in \mathbb{R}^2 \), representing the longitude and latitude. The goal is to develop a function \( f_t: \mathbb{R}^d \rightarrow \mathbb{R}^2 \) that generalizes well to the target domain.

We denote the source model as a DNN parameterized by \( \theta_s \), where \( \theta_s \) represents the set of learned weights obtained from training on the source domain. The corresponding function is defined as:
\begin{equation}
    f_s(x_i; \theta_s) = \hat{y}_i,
\end{equation}
where \( \hat{y}_i \) is the predicted output for input \( x_i \) in the source domain.

To adapt the model to the target domain, the target model parameters \( \theta_t \) are initialized with the source parameters \( \theta_s \) and fine-tuned using the target domain data. The resulting function is:
\begin{equation}
    f_t(x_j; \theta_t) = \hat{y}_j,
\end{equation}
where \( \theta_t \) denotes the updated weights optimized for the target domain.

\subsubsection{Loss Design}
To enable the model to balance knowledge transfer between the source and target domains, we employ a combined loss function that integrates both the source domain loss \( \mathcal{L}_s \) and the target domain loss \( \mathcal{L}_t \). The total loss is formulated as:
\begin{equation}
    \mathcal{L}_{total} = \lambda_s \mathcal{L}_s(\theta_s) + \lambda_t \mathcal{L}_t(\theta_t),
\end{equation}
where \( \lambda_s \) and \( \lambda_t \) are the respective weighting factors, optimized through cross-validation to enhance performance on the target task. This setup allows the model to adapt to the target while retaining useful source features. In other words, the total loss jointly guides the model to preserve prior knowledge while adjusting to new environmental variations. The source and target domain losses (i.e., $\mathcal{L}_s$ and $\mathcal{L}_t$) are computed independently to capture errors within each respective dataset, using the Mean Squared Error (MSE) criterion:
\begin{equation}
    \mathcal{L}_s = \frac{1}{n_s} \sum_{i=1}^{n_s} \mathcal{L}(f_s(x_i; \theta_s), y_i),
\end{equation}
\vspace{-0.5cm}
\begin{equation}
    \mathcal{L}_t = \frac{1}{n_t} \sum_{j=1}^{n_t} \mathcal{L}(f_t(x_j; \theta_t), y_j),
\end{equation}
where \( \mathcal{L} \) represents the MSE, which is used to calculate the individual losses for both the source and target models. Here, \( f_s(\cdot; \theta_s) \) and \( f_t(\cdot; \theta_t) \) represent the source and target models parameterized by \( \theta_s \) and \( \theta_t \). This formulation ensures that both domains contribute proportionally to the learning objective, improving the model’s adaptability during transfer.

\subsubsection{Fine-Tuning}

During fine-tuning, the parameters of the target model \( \theta_t \) are initialized based on the parameters of the source model \( \theta_s \). The model then undergoes small adjustments \( \Delta \theta \) using the target domain data, which helps it adapt more effectively to the variations in the new environment while retaining valuable knowledge from the source task. This approach is expressed as:
\begin{equation}
    \theta_t = \theta_s + \Delta \theta,
\end{equation}
In simple terms, each parameter of the target model is the sum of the corresponding source parameter and a small update learned from the target data. These updates ensure a gradual adaptation, reducing the risk of performance degradation due to differences between the source and target domains.

While evaluating the Base, EV, and TL models provides useful insights into their individual strengths and adaptability in different environments, it is not sufficient to rely solely on accuracy-based comparisons. A thorough evaluation must also take into account other factors, including computational efficiency, data usage, and the practical advantages of TL. To support more comprehensive and effective  model comparison and deployment decision-making, we introduce a unified evaluation metric in the next subsection. This metric offers a systematic approach to balancing and quantifying these key elements, ensuring fair and meaningful comparisons across diverse model configurations.

\subsection{Unified Performance Metric Formulation}
\label{ME}

We introduced a comprehensive metric called the \textit{Model Efficiency (ME)} which integrates four key components: \textit{Performance Accuracy (PA)}, \textit{Energy Efficiency (EE)}, \textit{Data Efficiency (DE)}, and the \textit{Transfer Gain (TG)}. This unified metric enables a balanced comparison across different models more effectively, ensuring that improvements in one area do not compromise another. 
The ME metric for the $i$-th model, denoted as $ME_i$, is defined in (\ref{eq:ME}), where $w_{\text{PA}}, w_{\text{EE}}, w_{\text{DE}}, w_{\text{TG}}$ represent the weights assigned to each component. Here, \( i \in \{1, 2, \ldots, N\} \) indexes the evaluated models, where each index corresponds to the specific Base, EV, and TL model configurations—as used throughout Algorithm~\ref{alg:me_algo}.

\vspace{3cm}
\begin{strip}
\noindent\rule{\textwidth}{0.5pt} 
\begin{equation}
\label{eq:ME}
ME_i = (w_{{PA}} \times PA_i) + (w_{{EE}} \times EE_i) + (w_{{DE}} \times DE_i) + (w_{{TG}} \times TG_i),
\end{equation}
\end{strip}

In our study, we assign equal importance to all four components by setting  
$w_{\textnormal{PA}} = w_{\textnormal{EE}} = w_{\textnormal{DE}} = w_{\textnormal{TG}} = 0.25$. This uniform weighting ensures that each metric contributes fairly to the overall efficiency evaluation of the model. 

\medskip

\subsubsection{Performance Accuracy (PA)}

For each model, PA is calculated as:
\begin{equation}
\label{eq:PA}
    \mathrm{PA}_i = \left(1 - \frac{E_{\text{mean}, i}}{E_{\text{max}, i}}\right) \times 100,
\end{equation}

where \( E_{\text{mean}, i} \) is the mean localization error measured across all evaluation epochs during validation, and \( E_{\text{max}, i} \) is the maximum localization error representing the highest recorded value within the same evaluation process for the \( i^{\text{th}} \) model. A higher PA value indicates better model accuracy.


 
\medskip
\subsubsection{Energy Efficiency (EE)}

EE evaluates the computational cost of model training by quantifying energy consumption per unit of useful work accomplished. Let \(T_j^{(i)}\) be the duration of the \(j\)-th training epoch for the \(i\)-th model (in seconds), and let \(P_{\text{GPU}}(t)\) and \(P_{\text{CPU}}(t)\) be the instantaneous power consumption (in watts) of the GPU and CPU, respectively, at time \(t\). During training, power consumption is sampled at regular intervals. Let \(S_j^{(i)}\) be the total number of discrete power samples collected during the \(j\)-th epoch of the \(i\)-th model. The \(k\)-th power sample is denoted as:
\begin{equation}
P_{k,j}^{(i)} = P_{\text{GPU},k,j}^{(i)} + P_{\text{CPU},k,j}^{(i)}, \quad k = 1, \dots, S_j^{(i)},\quad j = 1,\dots, M,
\end{equation}

The average power consumption over the \(j\)-th epoch for the \(i\)-th model is then approximated as:
\begin{equation}
\overline{P}_j^{(i)} = \frac{1}{S_j^{(i)}} \sum_{k=1}^{S_j^{(i)}} P_{k,j}^{(i)} = \frac{1}{S_j^{(i)}} \sum_{k=1}^{S_j^{(i)}} \left( P_{\text{GPU},k,j}^{(i)} + P_{\text{CPU},k,j}^{(i)} \right),
\end{equation}

The total energy consumed in joules during the \(j\)-th epoch of the \(i\)-th model is:
\begin{equation}
E_j^{(i)} = \overline{P}_j^{(i)} \cdot T_j^{(i)},
\end{equation}

The overall energy efficiency of the \(i\)-th model, denoted as \(EE_i\), is computed as the average energy consumed per epoch across all \(M\) training epochs:
\begin{equation}
EE_i = \frac{1}{M} \sum_{j=1}^{M} E_j^{(i)},
\end{equation}

where \(M\) is the total number of training epochs. A higher \(EE_i\) value indicates a more energy-efficient training process for model \(i\), as it represents better performance or more computational work achieved per unit of energy consumed.

\medskip
\subsubsection{Data Efficiency (DE)}

DE quantifies the ability of a model to achieve strong performance using a reduced amount of data. It is defined as:
\begin{equation}
\label{eq:DE}
DE_i = \left(1 - \frac{D_i}{D_{\text{max}}} \right) \times 100,
\end{equation}
where \(D_i\) represents the percentage of data used to train model \(i\), and \(D_{\text{max}}\) denotes the maximum percentage of data available for training (in the given setup \(D_{\text{max}} = 100\%\)).
This metric rewards models that achieve competitive performance using less data. A higher \(DE_i\) value indicates better data efficiency, reflecting the model’s ability to generalize from smaller datasets.

\medskip
\subsubsection{Transfer Gain (TG)}

TG quantifies the relative performance improvement achieved through transfer learning (TL). It is defined as:
\begin{equation}
\label{eq:TG}
TG_i = \left(\frac{PA^{(i)}_{\text{TL}} - PA^{(i)}_{\text{EV}}}{PA^{(i)}_{\text{BM}}}\right) \times 100,
\end{equation}
where \(PA_{\text{TL}}^{(i)}\), \(PA_{\text{EV}}^{(i)}\), and \(PA_{\text{BM}}^{(i)}\) represent the performance accuracy of the TL, EV and base models, respectively, for the \(i^{\text{th}}\) model.
We consistently used the original dataset for the base model, while the EV and TL models were based on three noise-augmented datasets with \(\text{NF} \in \{2, 4, 8\}\).
A positive \(TG_i\) indicates that TL improves the performance of model \(i\) relative to both the EV and base models.

By integrating these four components into the unified ME metric, we achieve a comprehensive evaluation framework which 
ensures that improvements in one area do not compromise another, facilitating a fair and robust comparison across different models.

\section{Algorithm development and implementation}
\label{sec:algorithm}

\subsection{Pre-TL Model Development}
\label{pre-TL}
We begin by training a base model on the original dataset, which was collected during the initial measurement campaign, ensuring that it achieves satisfactory performance before proceeding to further steps. 
The training and validation procedures are formalized below, where the loss functions \(\mathcal{L}_{\mathrm{train}}\) and \(\mathcal{L}_{\mathrm{val}}\) measure the average localization error between predicted coordinates \(\hat{y}\) and true coordinates \(y\):

\begin{equation}
\mathcal{L}_{train} = \frac{1}{|\mathcal{D}_{train}|} \sum_{i=1}^{|\mathcal{D}_{train}|} | \hat{y}_i - y_i |, 
\end{equation}
\begin{equation}
\mathcal{L}_{val} = \frac{1}{|\mathcal{D}_{val}|} \sum_{m=1}^{|\mathcal{D}_{val}|} | \hat{y}_m - y_m |,
\end{equation}
where \(\mathcal{D}_{\mathrm{train}}\) and \(\mathcal{D}_{\mathrm{val}}\) are the training and validation datasets composed of signal-intensity/localization pairs, and \(\hat{y}_i, y_i, \hat{y}_m, y_m\) are the corresponding predicted and true coordinates. After the model shows convergence based on validation error, the final model weights \(W_j\) are saved. 

After training the base model and collecting benchmark metrics, we simulate environmental drift by injecting Gaussian noise into the training data. This generates multiple datasets representing environmental conditions at time \( t + i \), where \( i \) denotes the progression in time of data collection. The noise intensity is controlled by the standard deviation \(\sigma\) and scaled by a noise factor (NF). 
Each resulting model, trained on the specific noisy dataset, is referred to as the EV model.

\begin{equation}
\mathcal{D}_{train}^{noisy} = \mathcal{D}_{train} + \mathcal{N}(0, \sigma^2),
\label{eq:1}
\end{equation}
where \(\mathcal{N}(0, \sigma^2)\) is Gaussian noise with zero mean and variance \(\sigma^2\).
The standard deviation \(\sigma\) is scaled as follows:
\begin{equation}
\sigma = NF \cdot \sigma_{base},
\label{eq:2}
\end{equation}
where \(\sigma_{base}\) is the baseline standard deviation. 
The model is updated using Stochastic Gradient Descent (SGD), which updates the weights iteratively by minimizing the loss function:
\begin{equation}
W_{j+1} = W_j - \eta \nabla L(W_j),
\label{eq:3}
\end{equation}
where \(L(W_j)\) is the
loss function and \(\eta\) is the
learning rate. 

As a result, when each EV model is trained on its corresponding noise-injected dataset generated by applying a specific NF (2, 4, or 8), the localization error increases substantially compared to the base model’s performance on the original dataset. This performance decline underscores the need for employing TL to adapt the model to changing environmental conditions, ensuring the retention of its localization accuracy. 

\subsection{TL Algorithm Development and Implementation}



The TL algorithm as shown in Algorithm \ref{TL algorithm}, takes as input a set of received signal strength dataset denoted as $D_{train}$ and $D_{val}$, a noise variance \( \sigma^2 \), weighting factor \( \alpha \) which refers to a scalar hyperparameter used to balance multiple loss terms or regularization components in the objective function, and outputs the trained model weights \( W_j \) which refer to the trainable parameters of the neural network model. The server initializes the model with parameters \( W_j \) and sets the epoch counter \( j \).
The base model \( \mathcal{M}_{base} \) and its weights \( W_{base} \) are loaded, then transferred to the TL model \( \mathcal{M}_{TL} \), and the shared layer weights (layers reused from the base model are set accordingly. To facilitate TL, the shared layers of \( \mathcal{M}_{TL} \) are frozen, ensuring that these layers are not updated during subsequent training by setting their `trainable` attribute to `False`. During each training epoch, Gaussian noise with variance $\sigma^2$ is injected into $\mathcal{D}_{train}$ (original dataset), to simulate environmental variability, resulting in a noisy dataset $\mathcal{D}_{train}^{noisy}$. Fine-tuning is then performed using $\mathcal{D}_{train}^{noisy}$, where only the non-frozen (task-specific) layers are updated through backpropagation, leading to the updated weights $W_{r+1}$.

\begin{algorithm}
\scriptsize 
\caption{Transfer Learning for Localization}
\label{TL algorithm}
\SetAlgoLined
\SetKwInOut{Input}{Input}
\SetKwInOut{Output}{Output}

\Input{$\mathcal{D}_{train}$; /* User VLC dataset */}
\Input{$\mathcal{D}_{val}$; /* User VLC Dataset */}
\Input{$\alpha$; /* Output weights */}
\Input{$\sigma^2$; /* Variance of Gaussian noise */}
\Output{$W_j$; /* Trained model parameters */}

\BlankLine

\SetKwProg{ServerInit}{ServerInit}{}{}
\ServerInit{}{
    Initialize $W_j$, $j \gets 0$ \;
    /* Set initial weights and epoch counter */}

\SetKwProg{LoadBaseModel}{Load Base Model}{}{}
\LoadBaseModel{}{
    Load $\mathcal{M}_{base}$ and its weights $W_{base}$;
}

\SetKwProg{DefineTLModel}{Define TL Model}{}{}
\DefineTLModel{}{
    Define $\mathcal{M}_{TL}$;
}

\SetKwProg{TransferWeights}{Transfer Weights}{}{}
\TransferWeights{}{
    Transfer $W_{base}$ to $\mathcal{M}_{TL}$ and set shared layer weights;
}

\SetKwProg{FreezeSharedLayers}{Freeze Shared Layers}{}{}
\FreezeSharedLayers{}{
    \ForEach{layer in $\mathcal{M}_{TL}$}{
        \If{layer is in shared layers}{
            Freeze the layer by setting `layer.trainable = False;
            /* Preserve shared knowledge */}
        };
    }

\While{$j < M$ and not converged}{
    \SetKwProg{AddNoise}{Simulate EV Model 
    }{}{}
    \AddNoise{}{
        $\mathcal{D}_{train}^{noisy} = \mathcal{D}_{train} + \mathcal{N}(0, \sigma^2)$;
        /* Simulate environmental variations */
    }

    \SetKwProg{TrackEpochTime}{Track Epoch Time}{}{}
    \TrackEpochTime{}{
        $T_{start} \gets \text{start time}$\;
    }

    \SetKwProg{TrainModel}{Train Model}{}{}
    \TrainModel{}{
        Fine-tune the model with $\mathcal{D}_{train}^{noisy}$, updating only the non-frozen layers;\;
        Compute gradient:\;
        $\nabla L(W_j) = \frac{1}{|\mathcal{D}_{train}^{noisy}|} \sum_{i=1}^{|\mathcal{D}_{train}^{noisy}|} \nabla f(x_i, y_i, W_j)$\;
        Update weights:\;
        $W_{j+1} = W_j - \eta \nabla L(W_j)$ \;
        /* Gradient descent update on non-frozen layers */
    }

    \SetKwProg{TrackEpochTimeEnd}{Track Epoch Time End}{}{}
    \TrackEpochTimeEnd{}{
        $T_{end} \gets \text{end time}$\;
        $T_{j} = T_{end} - T_{start}$\;
    }

    \SetKwProg{EnergyConsumed}{Monitor Energy Consumed}{}{}
    \EnergyConsumed{}{
        $E_{j} = (P_{GPU} + P_{CPU}) \cdot T_{j}$\;
        $E_{total} += E_{j}$; /* Update cumulative energy */;
    }

    \SetKwProg{LocalizationError}{Compute Localization Error}{}{}
    \LocalizationError{}{
        $\mathcal{L}_{train} = \frac{1}{|\mathcal{D}_{train}|} \sum | \hat{y}_i - y_i |$;
        /* Avg. localization error */
        $\mathcal{L}_{val} = \frac{1}{|\mathcal{D}_{val}|} \sum | \hat{y}_i - y_i |$;
    }

    \SetKwProg{SuccessRate}{Compute Success Rate}{}{}
    \SuccessRate{}{
        $SR_{train} = \frac{|\mathcal{D}_{train}|_{\mathcal{L}_{train} < \delta}}{|\mathcal{D}_{train}|}$, 
        $SR_{val} = \frac{|\mathcal{D}_{val}|_{\mathcal{L}_{val} < \delta}}{|\mathcal{D}_{val}|}$;
    }

    \SetKwProg{SaveMetrics}{Save Metrics}{}{}
    \SaveMetrics{}{
        Save metrics: $\mathcal{L}_{train}$, $\mathcal{L}_{val}$, $SR_{train}$, $SR_{val}$, $E_{j}$, $T_{j}$;
    }

    $j \gets j + 1$\;
}

Save model: $transfer\_learning\_model.save.h5$\;
\end{algorithm}

As long as the maximum number of epochs $M$ has not been reached and the convergence condition is not met, the algorithm proceeds iteratively. For each epoch, the start and end times are recorded to compute the epoch duration $T_{\text{j}}$. The model is trained on $\mathcal{D}_{train}^{noisy}$ using gradient descent. The loss gradient $\nabla L(W_j)$ is calculated as the average gradient over the noisy dataset, and the model parameters are updated using the update rule shown in Eq.~\ref{eq:3}.

Energy consumption is monitored during training by calculating the energy used in each epoch, $E_{epoch} = (P_{\text{GPU}} + P_{\text{CPU}}) \cdot T_{\text{epoch}}$, and the cumulative energy consumption $E_{\text{total}}$ is updated accordingly. \hl{}Localization errors $\mathcal{L}_{train}$ and $\mathcal{L}_{val}$ are computed for the training and validation datasets, respectively, as the average absolute difference between the predicted and true localization coordinates. Success rates $SR_{train}$ and $SR_{val}$ are calculated as the proportion of samples for which the localization error does not exceed a predefined threshold $\delta$. In this work, we use $\delta = 1$~m, which reflects the minimum acceptable precision level for practical indoor localization scenarios in industrial environments. All performance metrics, including $\mathcal{L}_{train}$, $\mathcal{L}_{val}$, $SR_{train}$, $SR_{val}$, $E_{j}$, and $T_{\text{j}}$, are logged after each epoch. The process continues until either convergence is achieved or the maximum number of epochs is reached.

\subsection{ME Evaluation Algorithm }
Next, we evaluate the performance of our models by computing the ME, as shown in Algorithm \ref{alg:me_algo}. 
We design the ME algorithm to take as input the CSV files containing the performance metrics for the Base, EV, and TL models, as well as the precomputed TG values. The Base model data is loaded to compute the PA and EE using the function {extract\_metrics}. It initializes the lists for PA, EE, data sizes $D$, and TG values for the Base model. Since the Base model does not involve transfer learning, its TG value is set to zero by definition.

The algorithm then processes the EV and TL models. For each model, it extracts the PA and EE metrics. The data size $D$ is determined based on the model name (which represents different percentages of the dataset). If the model is a TL model, the corresponding TG value is retrieved from the TG table based on the data size and NF value. For the EV model, since it is trained directly on the noisy target data without any transferred knowledge, its TG value is not defined and is set to zero for consistency in the ME computation.

After collecting all PA, EE, and TG values, the DE is computed using (\ref{eq:DE}), and finally, the ME is calculated using (\ref{eq:ME}). The ME values are returned for further analysis, enabling comparative performance assessment of the different models.

\begin{algorithm}
\scriptsize
\caption{Model Efficiency Evaluation}
\label{alg:me_algo}
\SetAlgoLined
\SetKwInOut{Input}{Input}
\SetKwInOut{Output}{Output}

\Input{CSV files for Base, EV, TL models and TG}
\Output{ME values for each model}

\SetKwProg{Initialize}{Initialize}{}{}
\Initialize{}{ 
    Load Base model data\; 
    Compute $PA_{\text{base}}$, $EE_{\text{base}}$ using \texttt{extract\_metrics} $(model\_data)$\;
    $PA\_values \gets [PA_{\text{base}}]$\; 
    $EE\_values \gets [EE_{\text{base}}]$\; 
    $D\_values \gets [100]$; /* \% data for Base model */\;
    $TG\_values \gets [0]$; /* TG = 0 for Base model */\;
}

\SetKwProg{ProcessModels}{Process EV, TL Models}{}{}
\ProcessModels{}{ 
    \ForEach{model $(model\_name, model\_data)$ in EV, TL models list}{
        $(PA, EE) \gets \texttt{extract\_metrics}$;
        Determine data size $D$ from \texttt{model\_name} (e.g., 30\%, 50\%, 70\%, or 100\%);
        
        \eIf{\texttt{model\_name} contains `TL Model'}{
            \If{\texttt{model\_name} contains `NF2'}{
                $TG \gets$ Retrieve from TG table for $(D, NF = 2)$;
            }
            \ElseIf{\texttt{model\_name} contains `NF4'}{
                $TG \gets$ Retrieve from TG table for $(D, NF = 4)$;
            }
            \ElseIf{\texttt{model\_name} contains `NF8'}{
                $TG \gets$ Retrieve from TG table for $(D, NF = 8)$;
            }
        }{
            $TG \gets 0$; /* EV models get TG = 0 */
        }
        Append $PA$, $EE$, $D$, and $TG$ to $PA\_values$, $EE\_values$, $D\_values$, and $TG\_values$;
    }
}

\SetKwProg{ComputeDE}{Compute Data Efficiency (DE)}{}{}
\ComputeDE{}{
    Compute $DE$ using Eq. \eqref{eq:DE}\;
}

\BlankLine

\SetKwProg{ComputeME}{Compute Model Efficiency (ME)}{}{}
\ComputeME{}{ 
    Compute $ME$ using Eq. \eqref{eq:ME}\;
}

\SetKwProg{OutputResults}{Output Results}{}{}
\Output{}{ 
    Return the set of $ME$ values for further analysis;
}

\end{algorithm}
\section{Experimental Setup and Data Collection} 
\label{sec:setup}

To develop and evaluate the proposed VLC-based localization framework, a comprehensive experimental setup was implemented within a real-world industrial environment. This section details the design and specifications of the VLC transceiver system, the measurement setup deployed in the BOSCH factory, and the methodology followed for data collection. The objective is to capture signal strength data across various transmitter-receiver configurations under realistic factory conditions to support robust model training and validation.

\subsection{VLC Transceiver System}
\subsubsection{Transceiver Architecture}

\begin{figure}[!t]
    \centering
    \includegraphics[scale=1.2]{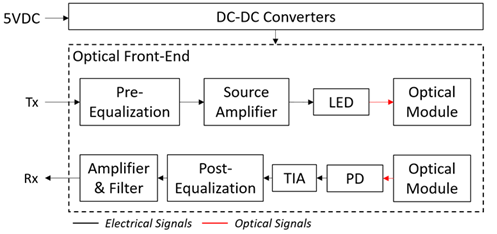}
    \caption{Architecture of the implemented VLC transceiver.}
    \label{fig:Architecture of the developed VLC transceiver}
\end{figure}

Fig. \ref{fig:Architecture of the developed VLC transceiver} represents the architecture of the implemented VLC transceiver. It comprises three primary components: the optical transmitter (TX), the optical receiver (RX), and the power supply unit. On the TX side, an input signal is provided via the 'Tx' interface. This signal undergoes pre-equalization to mitigate distortions introduced by the light source, followed by amplification to ensure compatibility with the electrical characteristics of the optical emitter. The modulated electrical signal is then converted into an optical signal by the light source. An optional optical module, such as a lens, may be employed to adjust the beam directivity depending on the measurement requirements.

After free space propagation, the optical signal sent by the TX of one VLC transceiver is collected by the RX of another VLC transceiver. This current signal is converted into a voltage signal and amplified by a transimpedance amplifier (TIA), then post-equalized to cope with the RX characteristic, and finally amplified again and filtered before being sent through the ‘Rx’ interface.

\subsubsection{Transceiver Specifications} 

TABLE~\ref{tab:parameters} summarizes the key specifications of the transmitter (TX), receiver (RX), and power supply units comprising the deployed VLC transceiver. Each unit is powered via a single USB-C female port requiring a 12 V, 1.2 A direct current (DC) source. The input voltage is regulated by a converter board that supplies a stable 5V DC to the optical front-end and signal processing circuits.

The transmitter employs a single LED driven by a current signal of 200 mA ± 100 mA, resulting in an optical output power ranging linearly from 250 mW to 750 mW. The emitted light exhibits a full-angle half-power beamwidth of 60°, with a central wavelength of 450 nm and an operational spectral range of 450–650 nm. The LED’s spectral width of 40 nm ensures operation entirely within the visible spectrum. Signal input to the TX chain is facilitated via an SubMiniature version A (SMA) connector to ensure robust coupling.

The receiver integrates multiple photodetectors connected in parallel, yielding an effective sensitive area (A\_Rx) of 600 mm². It is responsive across a broad spectral width of 400 nm, spanning the full visible range from 380 nm to 780 nm. The RX also exhibits a 60° full-angle half-sensitivity field of view, enhancing optical power collection from various angles. Like the transmitter, the RX chain employs an SMA interface for signal output.

\renewcommand{\arraystretch}{1.4}
\setlength{\tabcolsep}{3pt}
\begin{table}[!t]
\centering
\caption{VLC Transceiver Specifications.}
\begin{tabular}{|l|l|l|}
\hline
\textbf{Category} & \textbf{Parameter} & \textbf{Value} \\
\hline
\multirow{3}{*}{Power Supply} 
    & Supply voltage & 12 V \\
    & Supply current & 1.2 A \\
    & Power supply interface & USB-C female \\
\hline
\multirow{6}{*}{TX (Transmitter)} 
    & Driving current signal range & 200 mA $\pm$ 100 mA \\
    & Transmitted optical power range & 500 mW $\pm$ 250 mW \\
    & Full angle at half power & 60$^\circ$ \\
    & Wavelength of operation & 450 nm (450--650 nm) \\
    & Spectral width & 40 nm \\
    & TX chain interface & SMA \\
\hline
\multirow{5}{*}{RX (Receiver)} 
    & Effective sensitive area & 600 mm$^2$ \\
    & Photosensitivity width & 400 nm (380--780 nm) \\
    & Photosensitivity at 450 nm & 0.3 A/W \\
    & Full angle at half sensitivity & 60$^\circ$ \\
    & RX chain interface & SMA \\
\hline
\end{tabular}
\label{tab:parameters}
\end{table}

\subsection{VLC Measurement System}

\begin{figure}[!t]
    \centering
    \includegraphics[scale=0.2]{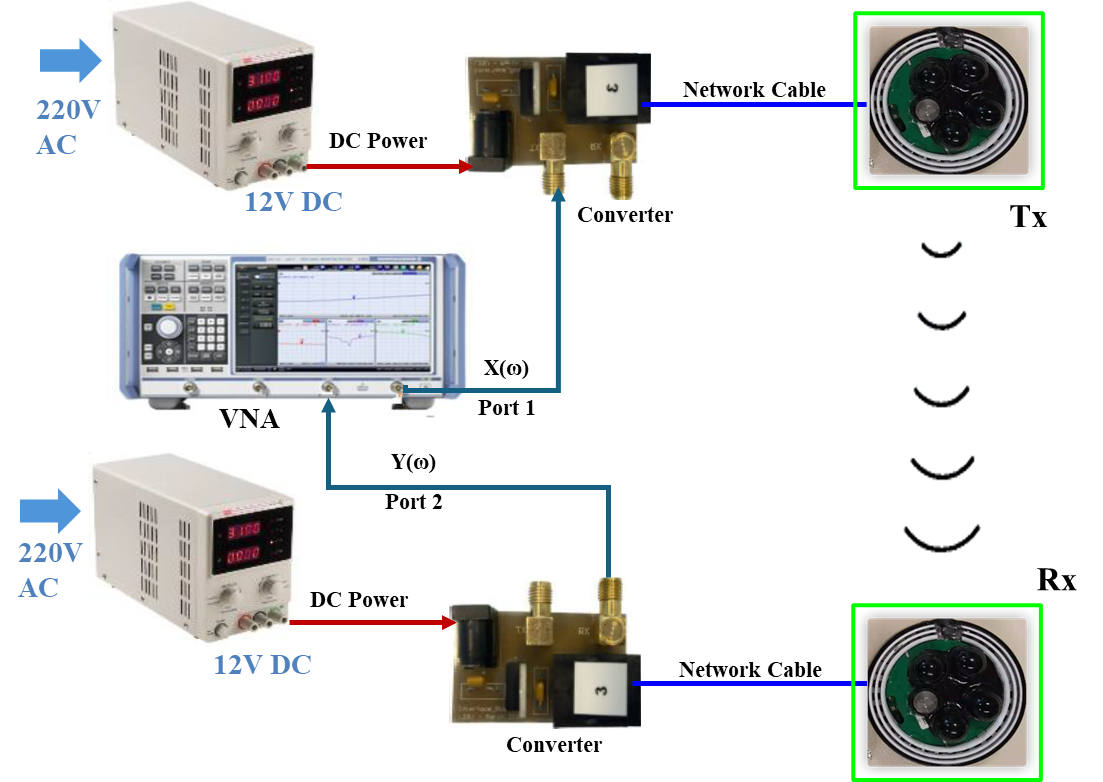}
    \caption{VLC measurement system.}
    \label{fig:vlc measurement setup}
\end{figure}

The VLC measurement setup, shown in Fig.~\ref{fig:vlc measurement setup}, was deployed at the BOSCH factory. It consists of two VLC transceivers, two converter boards, two DC power supplies, and a vector network analyzer (VNA). Each converter board combines the RF signal from the VNA with DC power, enabling simultaneous signal transmission and device powering through a single connection. During measurements, the VNA generates a frequency signal that modulates the optical carrier. The corresponding optical signal is received and converted by the RX unit, and the resulting electrical signal is captured and recorded by the VNA. This configuration enables high-precision measurement of the received signal at various RX positions, providing the basis for training and evaluating the VLC-based localization model.

\subsection{Data Collection} 

\begin{figure*}[!t]
    \centering
    \includegraphics[scale=1.0]{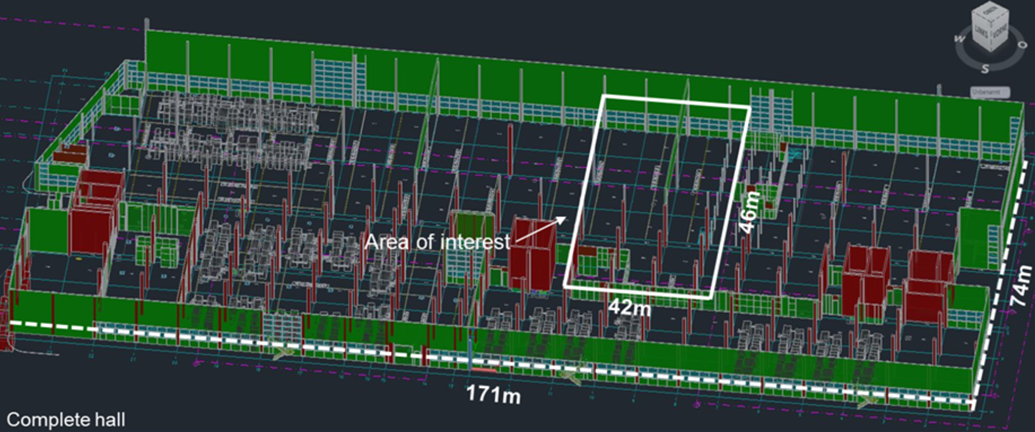}
    \caption{Schematic of the industrial hall at BOSCH in Blaichach.}
    \label{fig:hall}
\end{figure*}

We collected the data at the industrial hall of the BOSCH factory in Blaichach, Germany.
The complete hall has a dimension of 74 m $\times$ 171 m $\times$ 11 m, however the measurements were conducted in a portion of 46 m $\times$ 42 m $\times$ 11 m, as indicated in Fig. \ref{fig:hall}. This segment contains one production line and a large storage area. The limiting walls to the exterior are made from concrete, glass, and metal plates. In addition, concrete columns are mounted in a regular basis along the interior side. Multiple metallic structures such as tubes for ventilation and girders are found close to the ceiling.

\begin{figure*}[!t]
    \centering
    \begin{subfigure}[t]{0.45\textwidth}
        \centering
        \includegraphics[width=\linewidth]{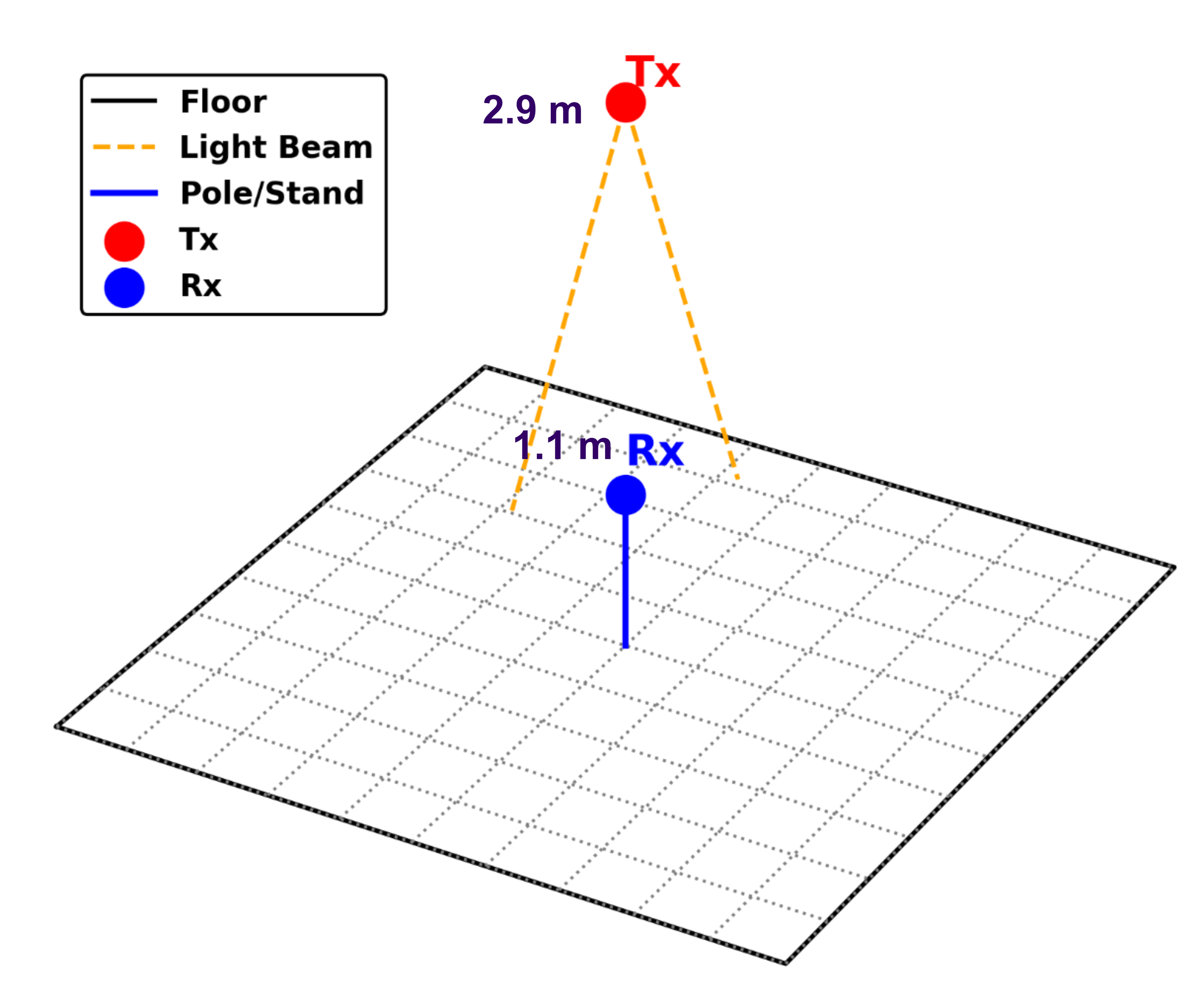}
        \caption{}
        \label{fig:txrx}
    \end{subfigure}
    \hfill
    \begin{subfigure}[t]{0.45\textwidth}
        \centering
        \includegraphics[width=\linewidth]{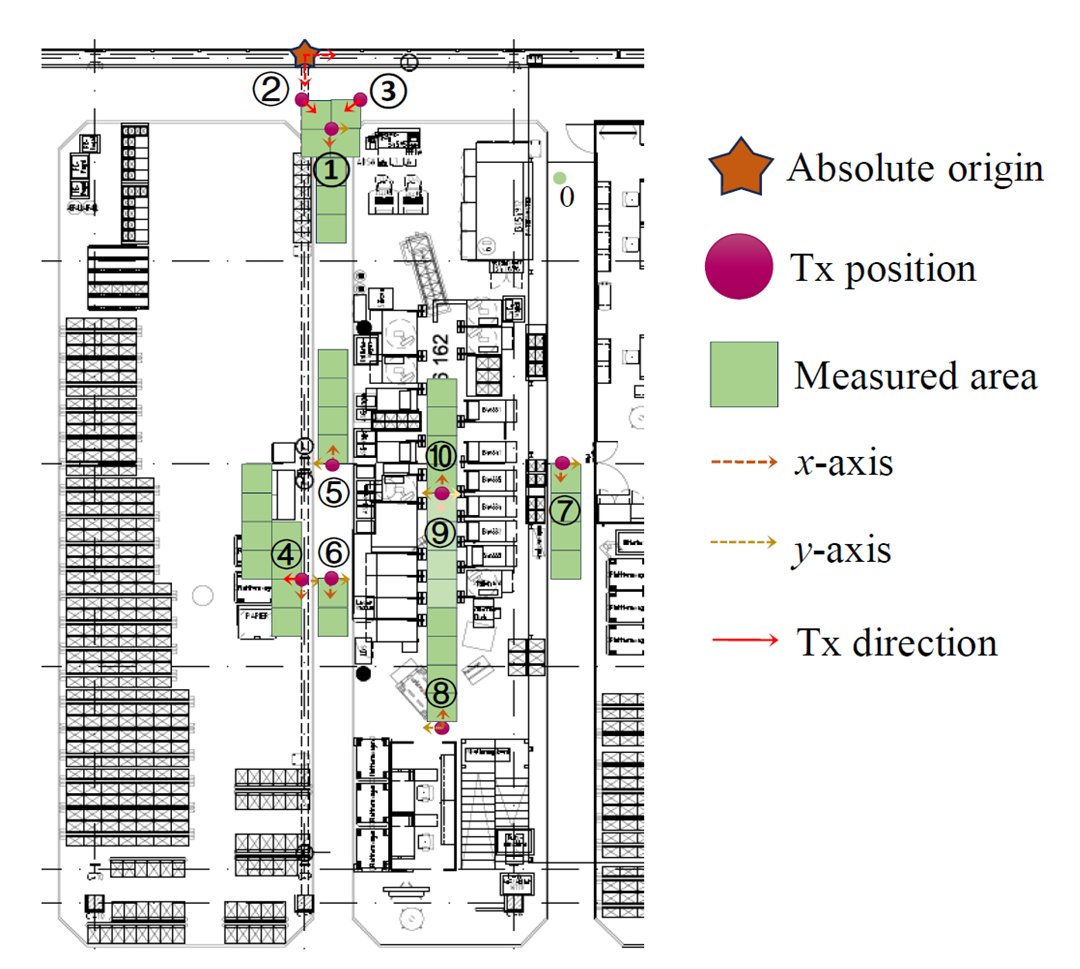}
        \caption{}
        \label{fig:floor_plan}
    \end{subfigure}
    \caption{Illustrations of (a) Tx/Rx setup, and (b) Floor plan layout.}
    \label{fig:txrx_floor}
\end{figure*}

Fig. \ref{fig:txrx} illustrates the overview of the transceiver setup and the grid layout used in the experiment. Taking into account the real-world layout and operational structure of the production line, the heights of the transmitter and receiver front-ends were fixed at 2.9 m and 1.1 m, respectively. 
For consistent analysis and simplified statistical processing, the receiving area was divided into uniform grids of 0.2 m $\times$ 0.2 m. The intersections of these grids served as the designated measurement points where data was collected.
Fig. \ref{fig:floor_plan} displays the actual positions of the transmitters and the measured area within the production line, as marked on the factory-provided floor plan. The orange star represents the absolute reference origin used in the data collection process. Based on the practical constraints of the production environment, 10 transmitter measurement areas were selected, with an average of approximately 160 receiver positions sampled at each location. At each of these positions, signal measurements were recorded continuously over a 5-second period to capture temporal variations caused by environmental noise and system fluctuations. This sampling duration strikes a balance between capturing dynamic effects and ensuring measurement efficiency. The average of the collected samples was then computed to yield a single Received Signal Strength Indicator (RSSI) value per position, thereby reducing the influence of transient anomalies and ensuring a stable representation of the signal quality for localization purposes. In this work we treat the initially measured dataset as the ground truth for the Base Model. The measurements were collected on-site in the operational production area, capturing realistic ambient lighting, machinery interference, and other environment-specific noise sources. Thus, the Base dataset is not a synthetic or noise-free reference but an empirical baseline that reflects actual factory conditions. This dataset serves as the ground-truth reference against which the EV and TL models were compared. The 3D coordinates and located areas of all transmitter positions are detailed in TABLE~\ref{tx positions}.

\begin{table}[!t]
\centering
\caption{3D Coordinates and Located Areas of Transmitters.}
\label{tx positions}
\renewcommand{\arraystretch}{1.3}
\begin{tabular}{|c|c|p{4cm}|}
\hline
\textbf{Transmitter ID} & \textbf{Position (x, y, z)} & \textbf{Located Area} \\
\hline
TX1 & (2.6, 0.8, 2.9) & Near control station and transparent windows \\
\hline
TX2 & (1.7, -0.2, 2.9) & Near control station and transparent windows \\
\hline
TX3 & (3.7, 1.8, 2.9) & Near control station and transparent windows \\
\hline
TX4 & (17.1, 0.2, 2.9) & Material storage zone \\
\hline
TX5 & (12.9, 0.8, 2.9) & Rear of production machinery (beside cement pillars) \\
\hline
TX6 & (17.1, 0.8, 2.9) & Rear of production machinery (adjacent to storage) \\
\hline
TX7 & (14.2, 8.5, 2.9) & Beneath conveyor belts near lockers \\
\hline
TX8 & (23.2, 4.6, 2.9) & Main production line \\
\hline
TX9 & (15.2, 4.6, 2.9) & Main production line \\
\hline
TX10 & (13.5, 3.6, 2.9) & Main production line \\
\hline
\end{tabular}
\end{table}



\section{Performance Evaluation and Analysis} 
\label{sec:results}
\subsection{TL Model Evaluation under different lighting conditions}

We evaluate the performance of the Base, EV, and the TL model. As previously mentioned in \ref{subsec:3A} and \ref{ME}, the base model is always trained using the original collected dataset while the EV model is trained on noise-injected datasets, where NF $\in \{2, 4, 8\}$ to simulate different lighting conditions. The TL model, on the other hand, is designed to handle such noisy datasets more effectively by leveraging knowledge from a previously trained base model. Its primary goal is to enhance performance metrics under varying environmental conditions. The DNN model shared across all three configurations, consists of five hidden layers. The ADAM optimizer is applied with a learning rate of 0.0001, and the Mean Square Error (MSE) loss function is used. 
We note that 80\% of the data samples were allocated for training, while the remaining 20\% were held out exclusively for validation across all experimental conditions (Base, EV, and TL models). For TL experiments, the model was first pre-trained on the source domain (Base dataset), then fine-tuned using varying portions (30\%, 50\%, 70\%, or 100\%) of the training subset from the target domain (with NF = 2, 4, or 8), and finally evaluated on the corresponding held-out validation subset from the target domain. This separation ensures that all reported performance metrics represent the model's true generalization capability on unseen samples. Training is conducted for 600 epochs with a batch size of 64. To account for variability due to random initialization and stochastic optimization, each experiment was repeated multiple times with different random seeds. Reported performance metrics (accuracy, EE, and ME) represent the average values across these runs with standard deviations reported to quantify uncertainty. A summary of all the parameters used in these experiments is provided in TABLE~\ref{Table:TL simulation settings}.


\begin{table}[!t]
\caption{TL Simulation Settings.}
    \centering
    \setlength{\tabcolsep}{11pt} 
    \renewcommand{\arraystretch}{1.3}
    \begin{tabular}{l l c}
        \specialrule{1.2pt}{0pt}{0pt} 
        \textbf{Parameter} & \textbf{Description} & \textbf{Value} \\
        \specialrule{0.8pt}{0pt}{0pt} 
        $d$ & Input dimension & 10 \\ \hline
        Optimizer & Model optimizer & ADAM \\ \hline
        $\eta$ & learning rate & 0.0001 \\ \hline
        $B$ & Batch size & 64 \\ \hline
        $E$ & Number of epochs per iteration & 600 \\ \hline
        $L$ & Number of hidden layers & 5 \\ \hline
        $h$ & Hidden layer size & 512 \\ 
        \specialrule{1.2pt}{0pt}{0pt} 
    \end{tabular}
    \label{Table:TL simulation settings}
\end{table}

\begin{figure*}[!t]
    \centering
    \begin{minipage}{0.32\textwidth}
        \centering
        \includegraphics[width=\linewidth]{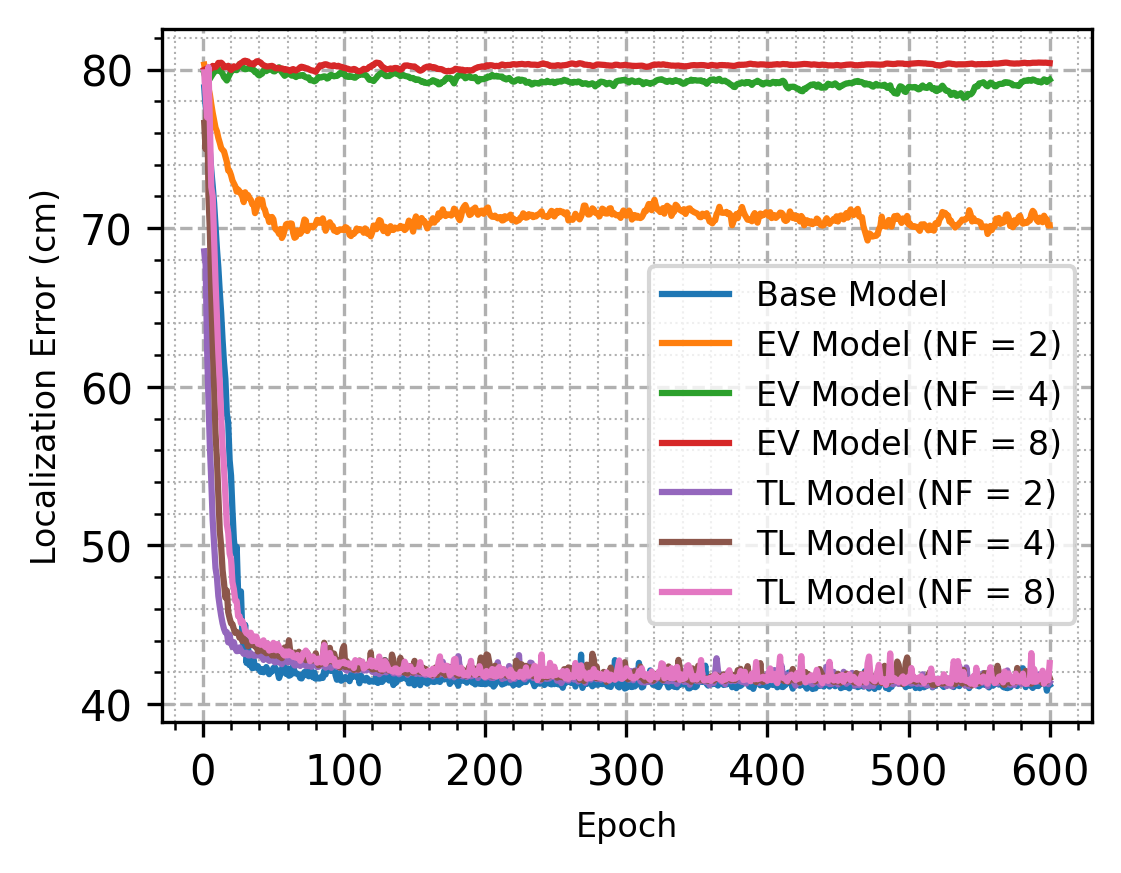}
        \subcaption{}
        \label{fig:loc_error}
    \end{minipage}
    \hfill
    \begin{minipage}{0.32\textwidth}
        \centering
        \includegraphics[width=\linewidth]{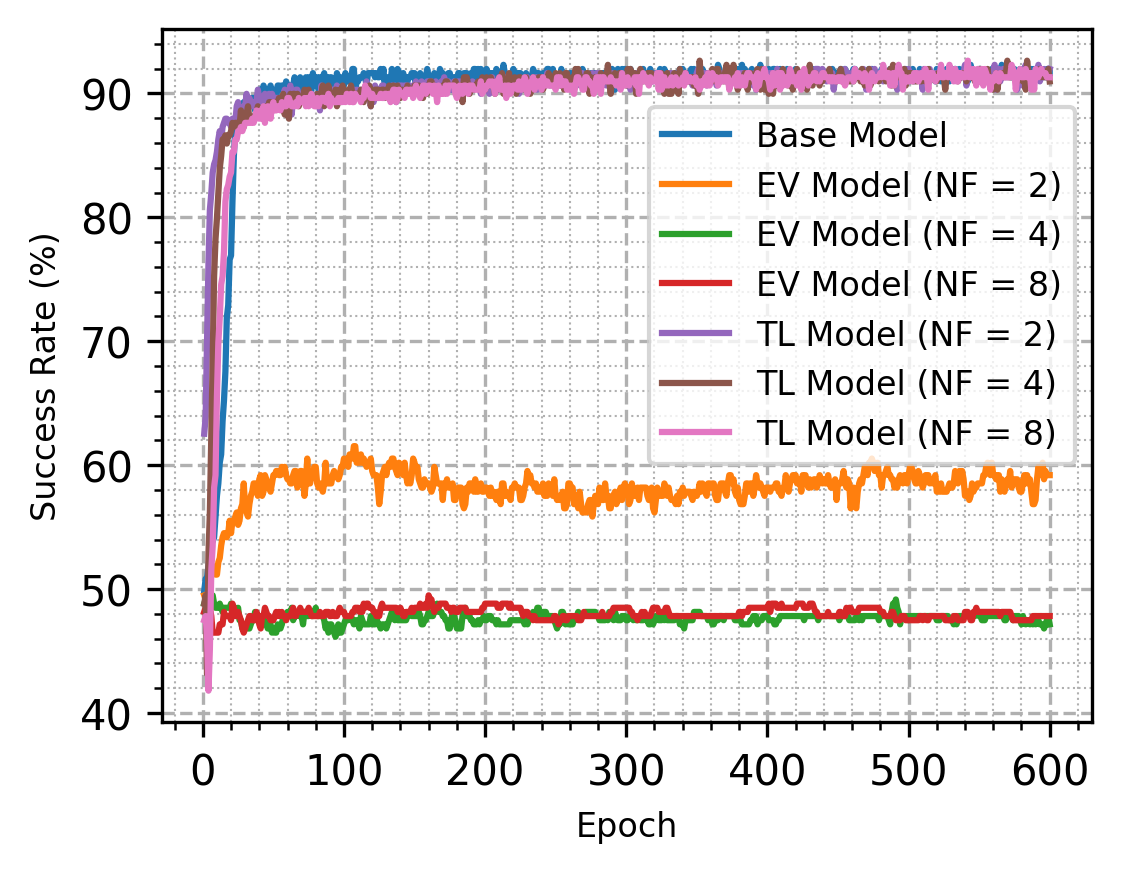}
        \subcaption{}
        \label{fig:success_rate}
    \end{minipage}
    \hfill
    \begin{minipage}{0.32\textwidth}
        \centering
        \includegraphics[width=\linewidth]{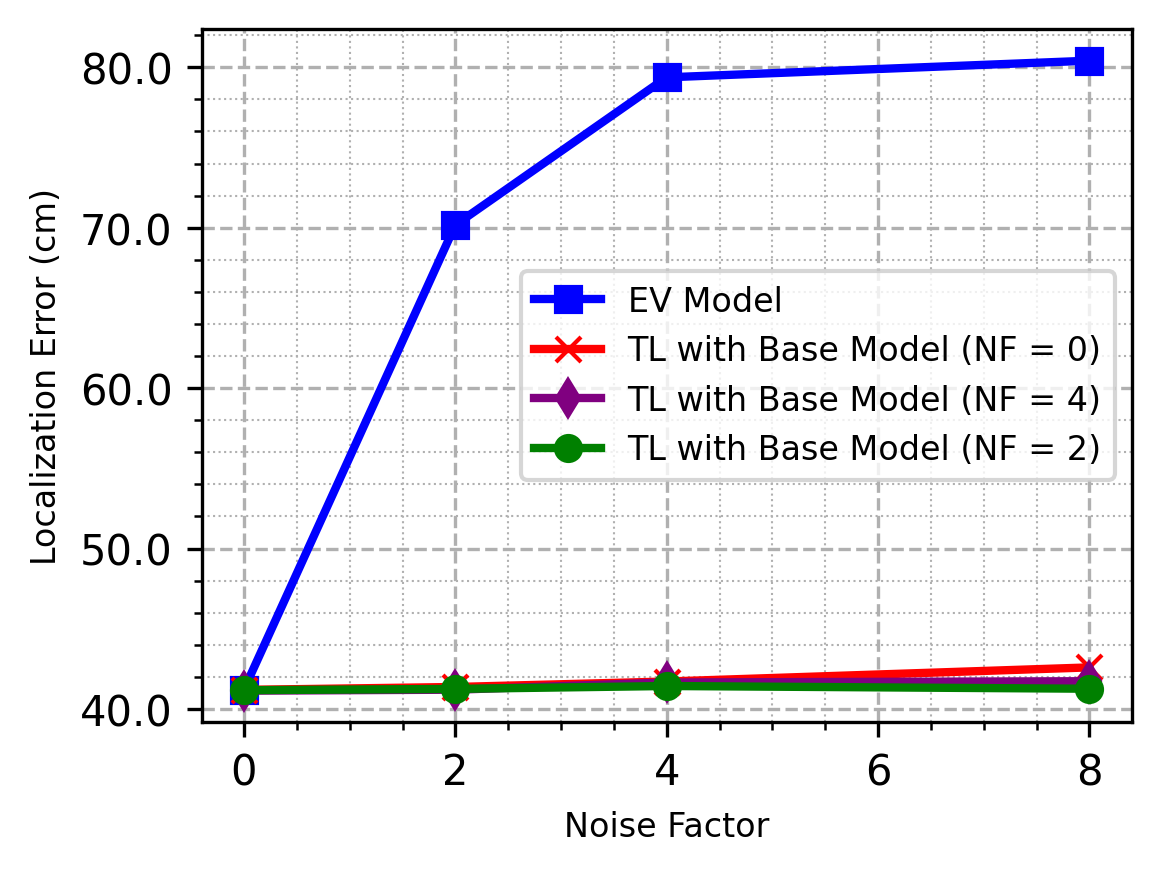}
        \subcaption{}
        \label{fig:learning_curve}
    \end{minipage}    
    \caption{TL performance evaluation: (a) Localization Error (b) Success Rate (c) TL of Base Model with different NFs.}
    \label{fig:performance_eval}
\end{figure*}

\begin{table*}[!t] 
\centering
\caption{Comparison of final metrics across models with different NFs, using the base model at NF = 0.}
\label{tab:final_metrics_combined}
\begin{tabular}{|l|c|c|c|c|c|c|c|c|}
\hline
\textbf{Final Metric Value} & \textbf{NF = 0} & \multicolumn{2}{c|}{\textbf{NF = 2}} & \multicolumn{2}{c|}{\textbf{NF = 4}} & \multicolumn{2}{c|}{\textbf{NF = 8}} \\
\hline
\textbf{{Training Model}} & \textbf{Base Model} & \textbf{EV Model} & \textbf{TL Model} & \textbf{EV Model} & \textbf{TL Model} & \textbf{EV Model} & \textbf{TL Model} \\
\hline
\text{Training Localization Error (cm)} & 41.13 & 70.38 & 41.37 & 78.99 & 41.74 & 79.91 & 42.42 \\
\hline
\text{Validation Localization Error (cm)} & 41.21 & 70.17 & 41.37 & 79.37 & 41.71 & 80.41 & 42.59 \\
\hline
\text{Training Success Rate (\%)} & 93.07 & 57.76 & 93.07 & 47.25 & 92.32 & 47.75 & 92.49 \\
\hline
\text{Validation Success Rate (\%)} & 91.64 & 59.20 & 91.97 & 47.16 & 90.97 & 47.83 & 91.30 \\
\hline
\text{Cumulative Energy Consumed (J)} & 852.79 & 1571.35 & 1064.91 & 1923.38 & 1082.56 & 976.70 & 867.22 \\
\hline
\text{Cumulative Training Time (s)} & 93.04 & 196.03 & 124.00 & 214.88 & 126.17 & 123.08 & 92.06 \\
\hline
\text{Training Time per Epoch (s)} & 0.21 & 0.35 & 0.19 & 0.32 & 0.17 & 0.20 & 0.15 \\
\hline
\end{tabular}
\end{table*}

\begin{table*}[!t]
\caption{Localization error comparison of TL models at different NFs.}
\centering
\renewcommand{\arraystretch}{1.0}
\begin{tabular}{|c|c|c|c|c|}
\hline
\textbf{NF} & \textbf{EV Model (cm)} & \textbf{TL as Base Model (cm)} & \textbf{TL as (EV Model NF =2) (cm)} & \textbf{TL as (EV Model NF =4) (cm)} \\
\hline
0 & - & 41.19 & 41.17 & 41.14 \\
\hline
2 & 70.17 & 41.37 & 41.25 & 41.20 \\
\hline
4 & 79.37 & 41.71 & 41.43 & 41.65 \\
\hline
8 & 80.41 & 42.59 & 41.26 & 41.73 \\
\hline
\end{tabular}
\captionsetup{justification=centering}
\label{tab:localization_error}
\end{table*}

Fig. \ref{fig:loc_error} illustrates the localization error observed during the validation phase. According to the results presented in TABLE~\ref{tab:final_metrics_combined}, the base model attains the best performance compared to the EV model, exhibiting a localization error of 41.21 cm. In contrast, the EV model showed considerable vulnerability to noise: its error escalates to 70.17 cm at $NF=2$, and further to 80.41\,cm at $NF=8$, corresponding to a 95.14\% degradation relative to the base model. On the other hand, the TL model effectively resists noise interference, maintaining low errors of 41.37 cm and 42.59 cm at $NF = 2$ and $NF = 8$ respectively, reflecting only minor increases of 0.39\% and 3.35\% over the base model. The same trend is observed in the success rate shown in Fig.~\ref{fig:success_rate}, reinforcing the superior noise resilience of the TL model compared to the EV approach.

While our localization error may not surpass some state-of-the-art methods, our principal aim is to introduce a TL framework that optimally balances accuracy, noise immunity, and minimal data requirements in an energy-efficient manner. Extending this TL strategy to more complex architectures beyond DNNs could further decrease the error at the expense of higher computational and power costs.


To further assess the TL model's robustness, we compared its performance when trained with different base models, as presented in Fig. \ref{fig:learning_curve}. TABLE~\ref{tab:localization_error} shows that using a base model trained with $NF = 2$ yields a slight error improvement of 0.29\% over the TL model trained with a $NF = 0$. Similarly, for $NF = 4$, the error is reduced by 0.67\% compared to the baseline TL model. These findings suggest that while the choice of a base model trained with noise does slightly improve localization accuracy, the effect is minimal, indicating low sensitivity to the base model selection across different noise levels.
Overall, although localization precision degrades as noise intensifies, the TL models mitigate this effect and enhance performance stability.

\subsection{TL model evaluation with limited data}

\begin{figure}[!t]
    \centering
    \includegraphics[scale=0.85]{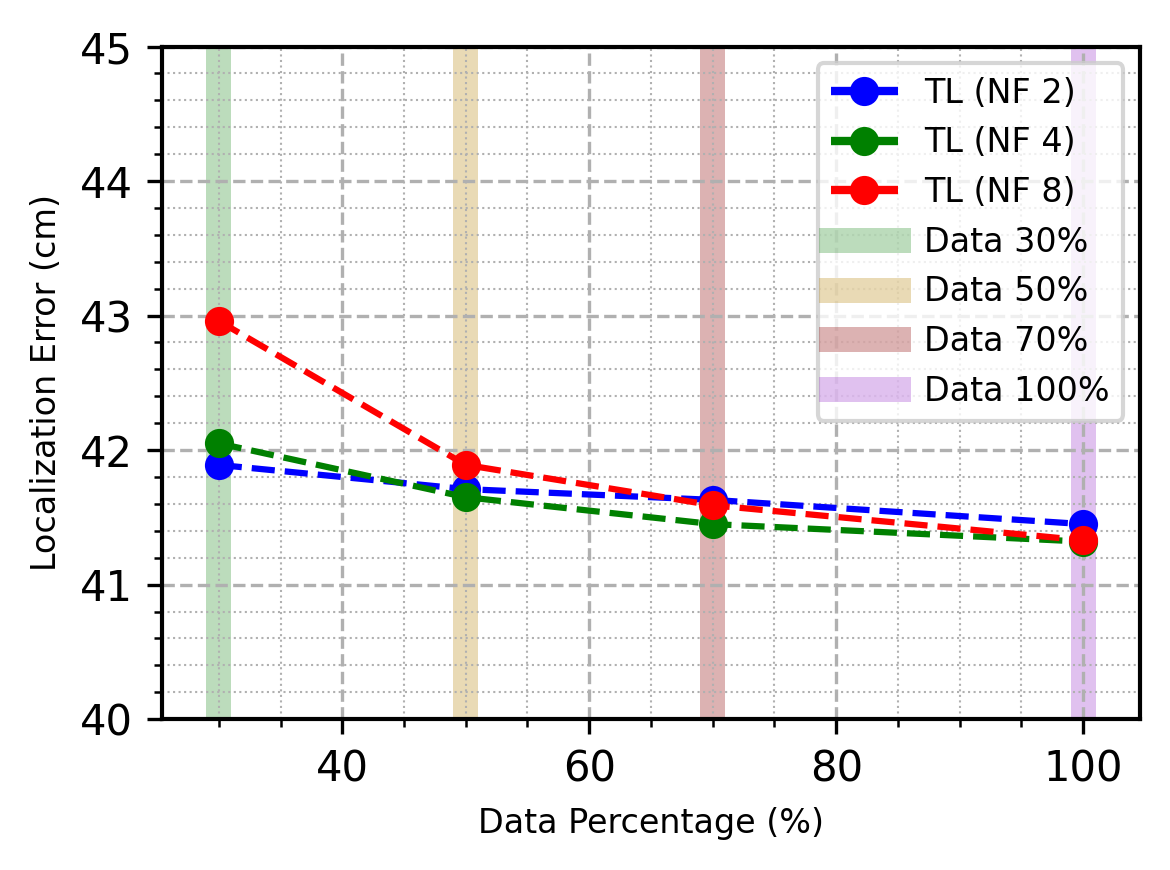}
    \caption{TL model evaluation with limited data.}
    \label{fig:TL limited data}
\end{figure}

We evaluate the performance of the TL model under varying data sizes and noise levels. As reported in TABLE~\ref{Table: datasize}, reducing the dataset to 30\% of its original size leads to a notable increase in localization error, reaching 86.78 cm. However, with the application of TL, the error significantly drops to around 42.96 cm, an improvement of over 45\%. This notable reduction demonstrates the TL model’s strong capability to maintain performance even when training data is limited. As illustrated in Fig. \ref{fig:TL limited data}, the localization error using just 30\% of the data, after applying TL, closely aligns with the error obtained when using the full dataset.

\begin{table}[!t]
\caption{Error values before and after TL at different NFs, comparing performance across various data sizes.}
\centering
\resizebox{\columnwidth}{!}{%
\begin{tabular}{|c|c|c|c|c|c|c|}
\hline
\textbf{Data Size} & \multicolumn{3}{c|}{\textbf{Error Before TL (cm)}} & \multicolumn{3}{c|}{\textbf{Error After TL (cm)}} \\
\hline
& \textbf{NF = 2} & \textbf{NF = 4} & \textbf{NF = 8} & \textbf{NF = 2} & \textbf{NF = 4} & \textbf{NF = 8} \\
\hline
30\% & 77.08 & 83.08 & 86.78 & 41.89 & 42.05 & 42.96 \\
\hline
50\% & 76.49 & 81.89 & 84.75 & 41.71 & 41.65 & 41.89 \\
\hline
70\% & 74.66 & 80.17 & 81.33 & 41.63 & 41.45 & 41.59 \\
\hline
100\% & 70.17 & 79.01 & 80.76 & 41.45 & 41.32 & 41.33 \\
\hline
\end{tabular}%
}
\label{Table: datasize}
\end{table}

\subsection{Key Metric Evaluation: TG, Accuracy and EE}

\begin{figure*}[!t]
    \centering
    \begin{minipage}{0.32\textwidth}
        \centering
        \includegraphics[width=\linewidth, height=4cm]{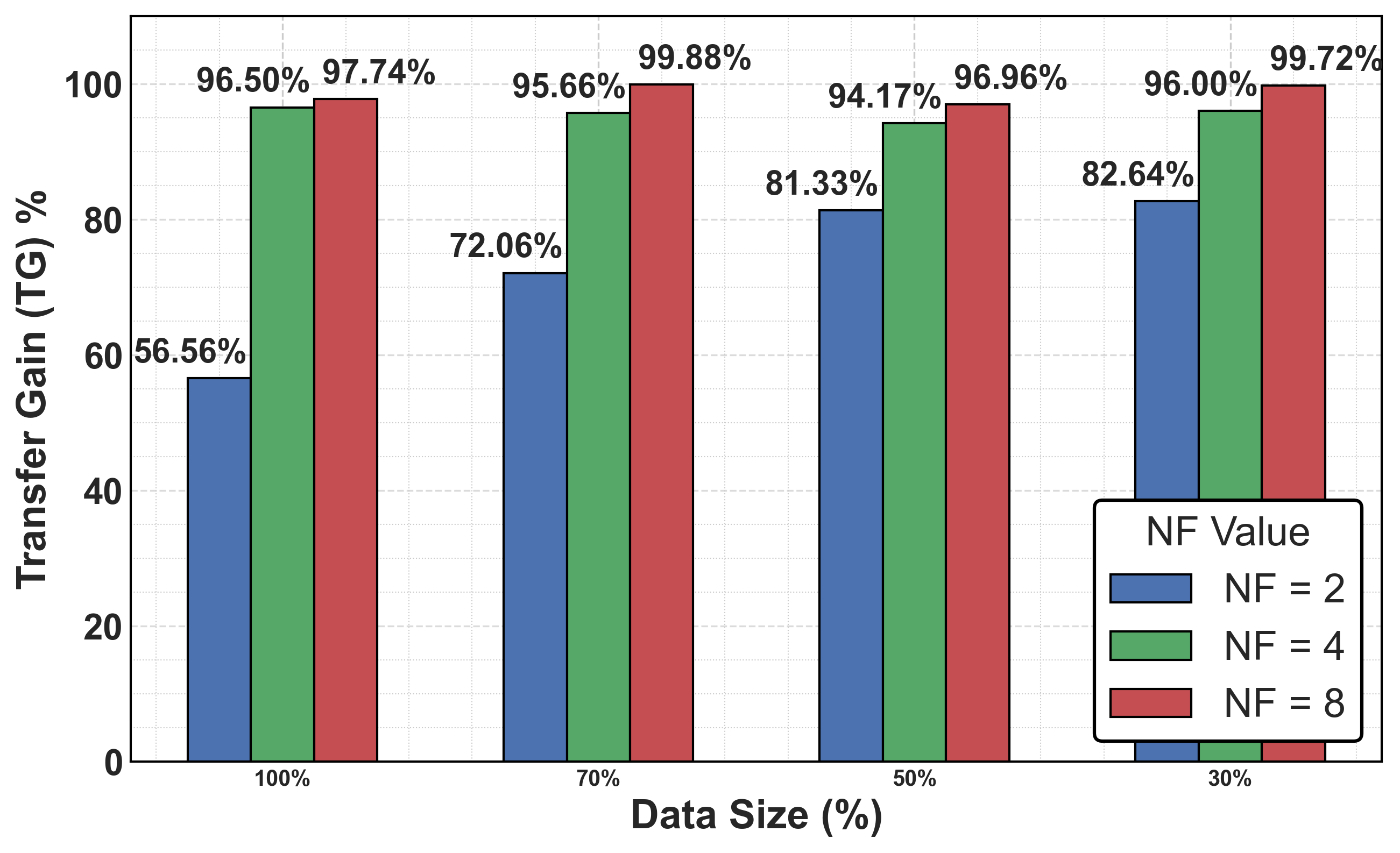}
        \subcaption{}
        \label{fig:TG_plot}
    \end{minipage}
    \hfill
    \begin{minipage}{0.32\textwidth}
        \centering
        \includegraphics[width=\linewidth, height=4cm]{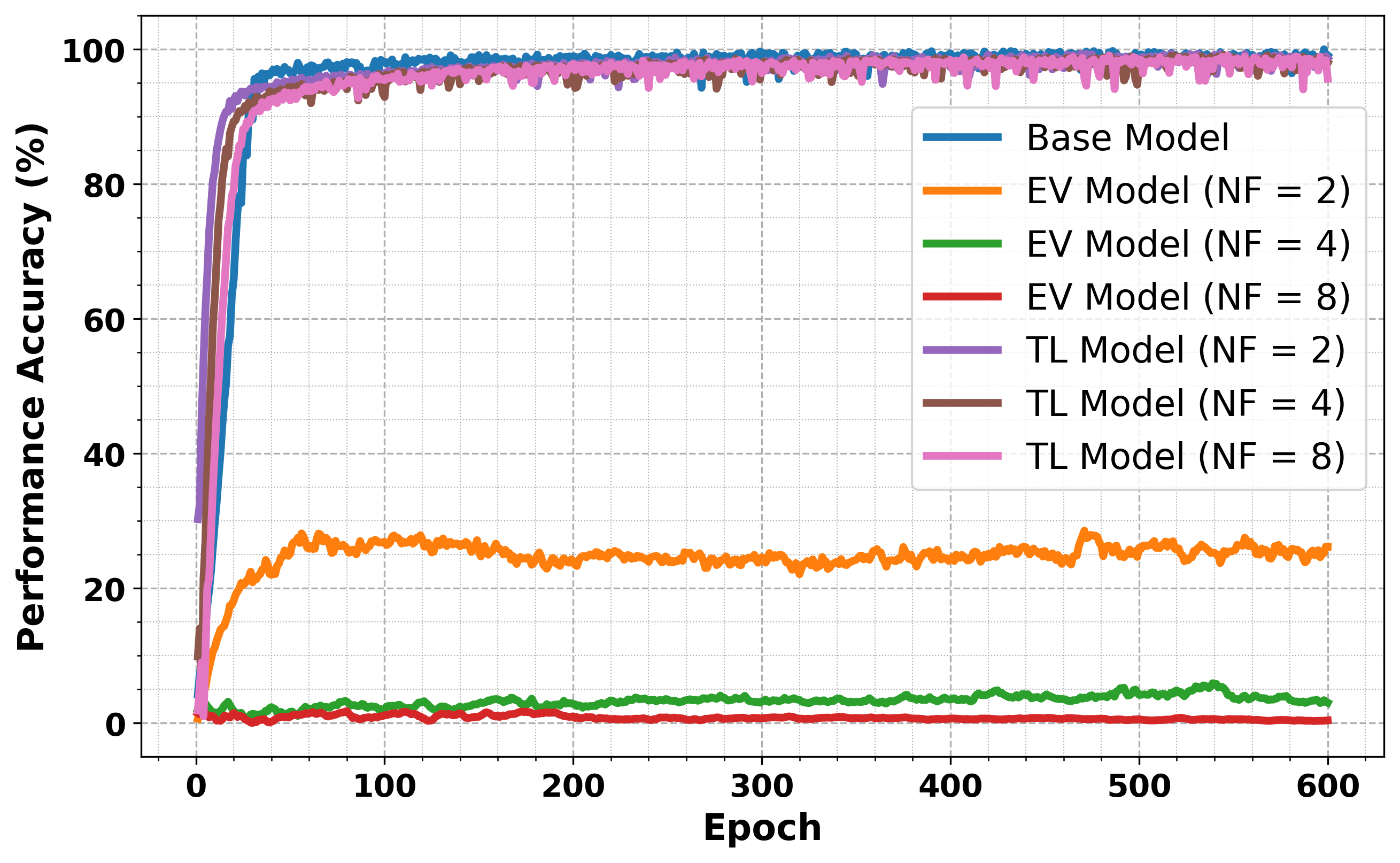}
        \subcaption{}
        \label{fig:PA_plot}
    \end{minipage}
    \hfill
    \begin{minipage}{0.32\textwidth}
        \centering
        \includegraphics[width=\linewidth, height=4cm]{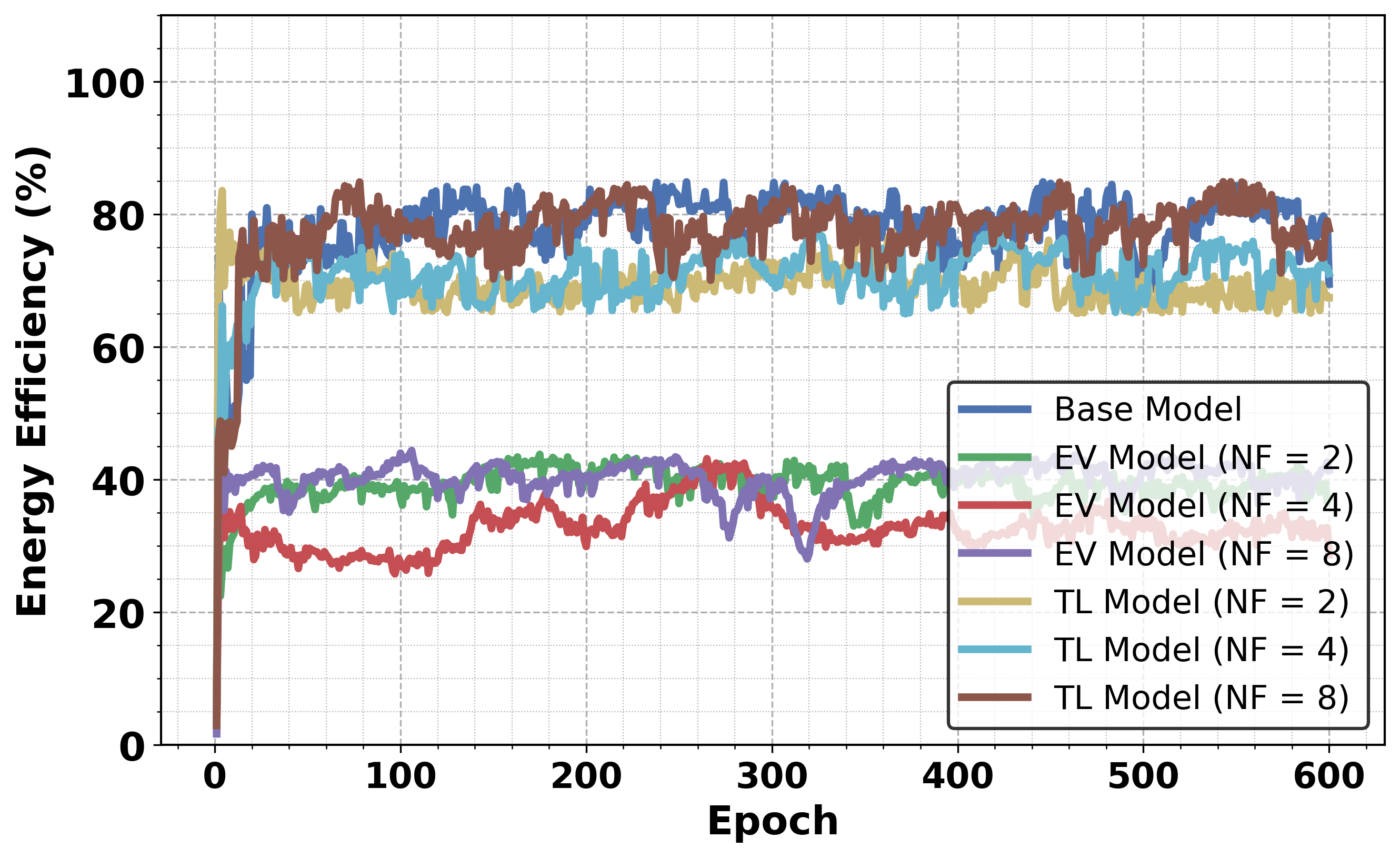}
        \subcaption{}
        \label{fig:EE_plot}
    \end{minipage}    
    \caption{Key Metrics Evaluation: (a) Transfer Gain for varying data sizes and NFs (b) Performance Accuracy of Base, EV, and TL models (c) Energy Efficiency trends derived from experimental measurements.}
    \label{fig:combined}
\end{figure*}

Fig.~\ref{fig:TG_plot} represents the TG for varying data sizes and noise levels. At 100\% data and NF = 2, the TG is comparatively low (56.56\%), due to a smaller accuracy gap between TL and EV models. As noise increases (NF = 4, 8), the EV model’s accuracy degrades more sharply, raising TG to 96.50\% and 97.74\%, respectively. With reduced data (70\%, 50\%, 30\%), the EV model performance drops further, while the TL model remains robust, widening the accuracy gap and elevating the TG. For example, at 70\% data and NF=8, TG peaks at 99.88\%, staying above 96.96\% even at 50\% and 30\%. Even for the lower noise setting of NF=2, the TG steadily increases from 56.56\% at 100\% data to 82.64\% at 30\% data. This confirms that TL is highly effective in scenarios where environmental variations and limited data availability are concerns.

Fig.~\ref{fig:PA_plot} highlights the PA of different models. The base model and TL models with varying noise levels i.e., $NF \in \{2, 4,8\}$ achieve high accuracy, nearing 100\% and maintaining stability throughout. However, the EV models exhibit significantly lower accuracy, with NF = 2 reaching around 25\% and with NF = 4 and NF = 8 remaining below 10\%. This suggests that while the base and TL models effectively generalize, the EV models struggle to adapt under environmental variations, particularly as noise factor increases. Overall, the results indicate that TL models provide a reliable solution for maintaining high accuracy even in the presence of environmental variations.  

\begin{figure*}[!t]
  \centering
  \includegraphics[width=\linewidth]{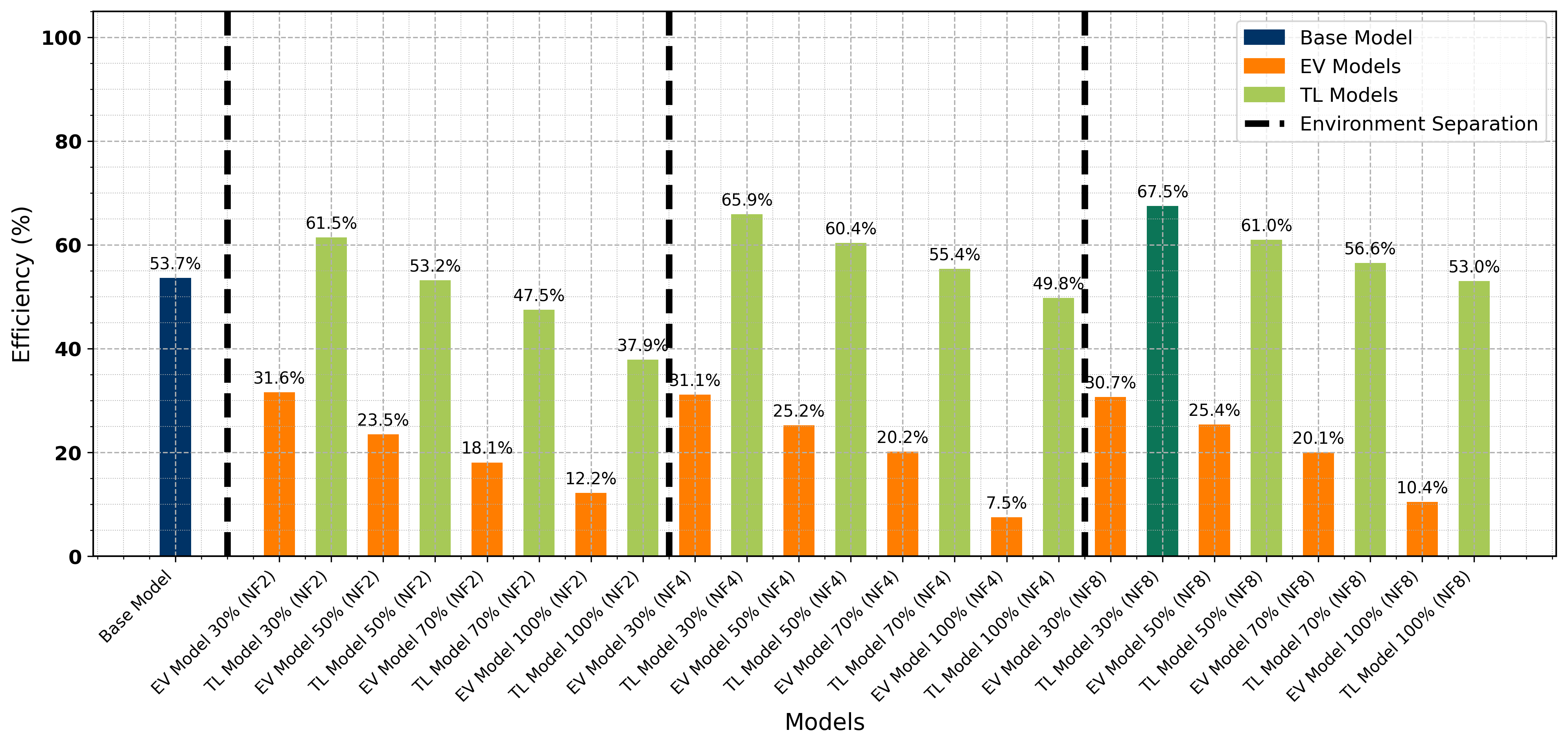}
  \caption{Model efficiency across different models and environmental conditions under varying data proportions and noise factors.}
  \label{fig:ME_plot}
\end{figure*}


Fig.~\ref{fig:EE_plot} highlights energy efficiency trends. The base model leads with ~80\% efficiency, followed closely by TL models, particularly TL (NF=2). In contrast, EV models are less efficient, averaging around 30\% for EV (NF=4) and around 40\% for EV (NF=8). This reflects the EV models’ instability under noise, leading to frequent adjustments, slower convergence, and higher power usage. TL models, however, maintain both accuracy and energy efficiency, demonstrating their robustness in noisy, variable environments.
All energy consumption values reported in this study were experimentally obtained using hardware-level monitoring during model training. GPU power usage was read directly from the NVIDIA driver through the \texttt{pynvml} library, and CPU utilization was recorded via \texttt{psutil}. For each epoch, instantaneous power values were sampled at regular intervals, averaged, and multiplied by the corresponding training duration to obtain the total energy consumed (in joules). These measurements were repeated across epochs and averaged to ensure consistency. The results therefore represent experimentally measured energy consumption values collected under identical hardware and runtime conditions.

It should be noted that the TL model exhibited very low variability (accuracy: 91.30 ± 0.51\%, EE: 72.10 ± 3.56), confirming excellent stability of the training process and reproducibility of the reported results. The EV model showed higher variability (accuracy: 47.83 ± 6.77\%, EE: 35.73 ± 6.48), which is expected due to the noise injection during training.

\subsection{Model Efficiency Evaluation} 

Fig.~\ref{fig:ME_plot} represents the ME trends across different models and environmental settings, with the ME computed through Eq.~\ref{eq:ME} for each scenario. 
The base model achieves 53.7\% ME and serves as a reference for comparison. Note that we keep the base model as a benchmark 
to show that, through TL the models trained on limited and noisy data can achieve higher ME than the original base model. 


The EV models trained under varying environmental conditions show a sharp drop in efficiency as NF increases, for example; EV Model 30\% (NF2) achieves 31.6\%, while EV Model 100\% (NF8) falls to just 10.4\%, highlighting their limited adaptability under environmental variations. 

However, TL models consistently outperform their EV counterparts, demonstrating robustness. TL Model 30\% (NF2) reaches 61.5\% efficiency, which is higher than the base model, showcasing the benefits of leveraging pre-trained knowledge. As NF increases, TL models continue to maintain strong efficiency levels, with TL Model 100\% (NF8) still achieving 53.0\% efficiency, significantly outperforming the corresponding EV models. 

These results emphasize ME as a reliable comparative metric in dynamic environments, as it captures multiple aspects of performance simultaneously. The results underscore the effectiveness of TL in sustaining efficiency across different environmental conditions, without the need for retraining from scratch.


\subsection{Generalization capacity of the best ME model}

\begin{figure*}[!t]
    \centering
    \begin{subfigure}{0.48\linewidth}
        \centering
        \includegraphics[width=\linewidth]{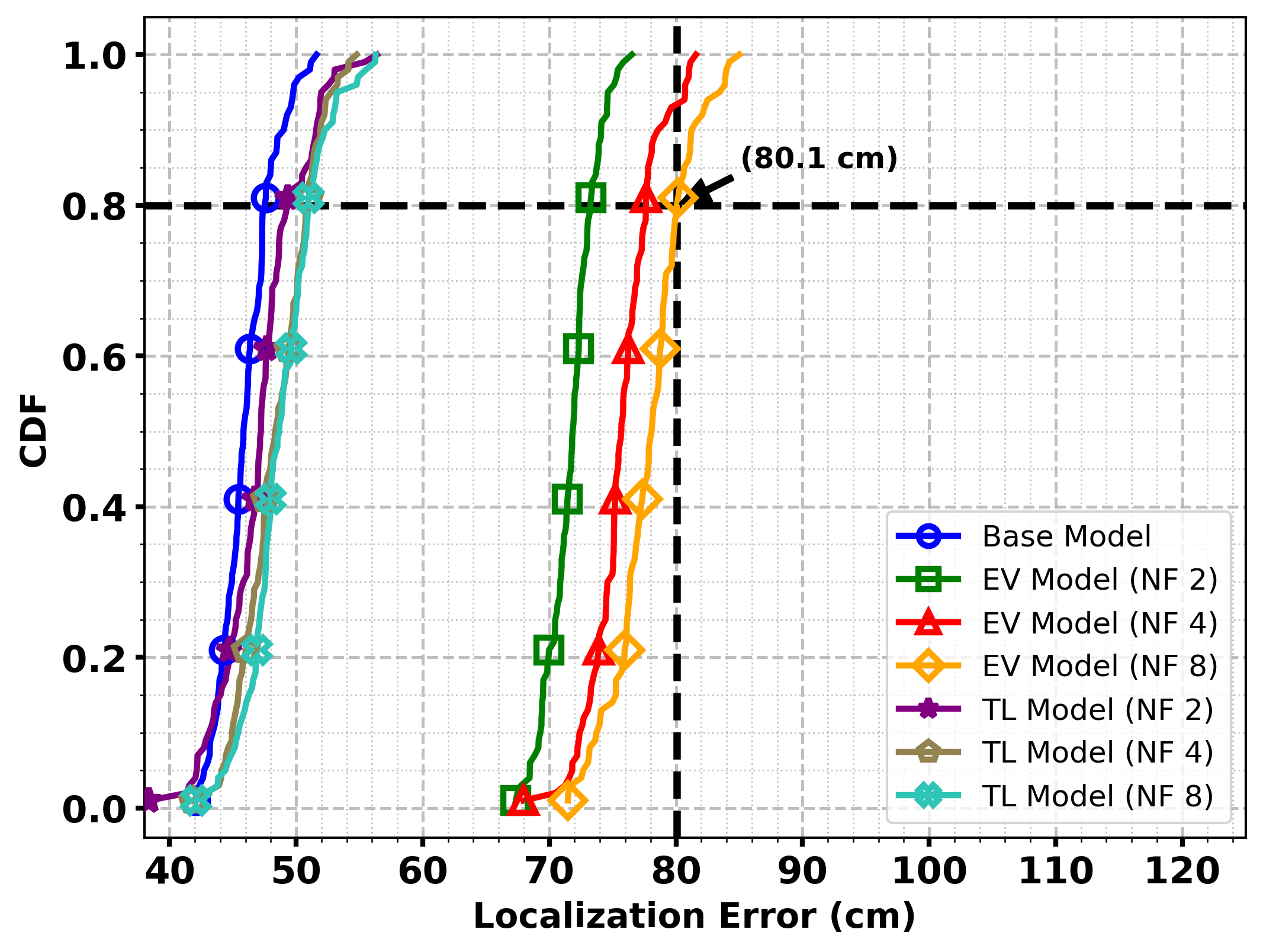}
        \caption{}
        \label{fig:cdf}
    \end{subfigure}
    \hfill
    \begin{subfigure}{0.48\linewidth}
        \centering
        \includegraphics[width=\linewidth]{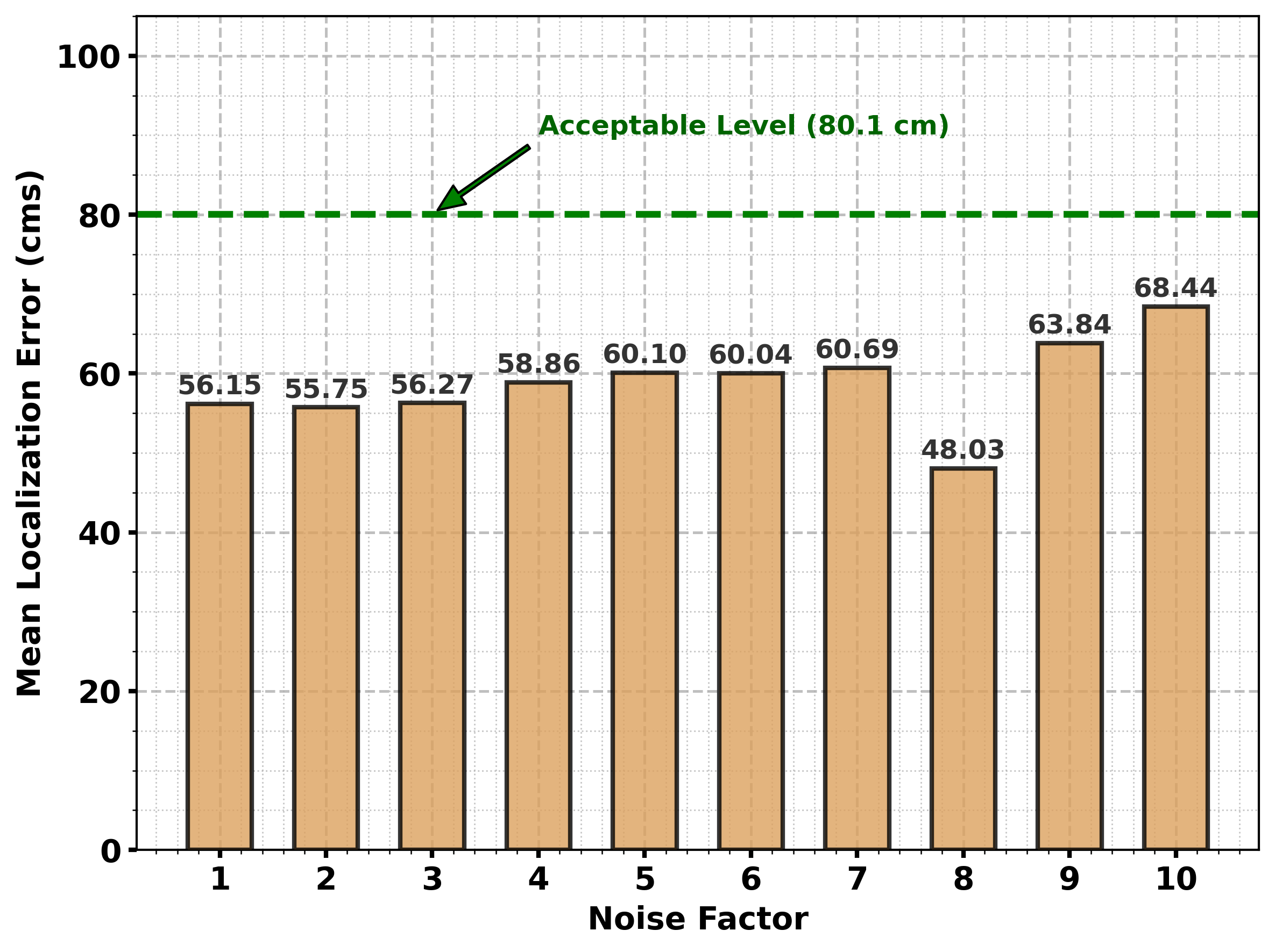}
        \caption{}
        \label{fig:best ME}
    \end{subfigure}
    \caption{(a) CDFs of localization error for Base, EV, and TL models under varying NFs and (b) Direct testing of the best ME model across different NFs. The dashed line indicates the 80th-percentile error threshold (80.1 cm).}

    \label{fig:combined}
\end{figure*}

To analyze the generalization ability of our best-performing TL model, we follow a two-step evaluation process. First, we define an acceptance threshold using the cumulative distribution functions (CDFs) of localization error, as shown in Fig.~\ref{fig:cdf}. These CDFs capture the error distribution for the base, EV, and TL models across varying NFs. We selected the 80th percentile of the TL model's error distribution as the threshold, which corresponds to approximately 80.1 cm (dashed line). This percentile is chosen because it reflects a balance between accuracy and robustness, ensuring that the model performs acceptably for the majority (80\%) of the test cases without being overly optimistic or overly strict.

Next, we evaluated the generalization capacity of the model that achieved the highest ME score, identified in Fig.~\ref{fig:ME_plot}. This model, trained at NF = 8, was directly tested across a broad range of environmental variations, from NF = 1 to NF = 10. As shown in Fig.~\ref{fig:best ME}, the MLE remains consistently below the predefined threshold of 80.1 cm across all noise conditions, demonstrating strong generalization capability. Notably, the lowest error occurs at NF = 8, which aligns with the training condition of the selected model, further validating its optimality under that specific setting.

Together, these figures serve complementary purposes: Fig.~\ref{fig:cdf} defines the error threshold based on the TL model’s distribution, while Fig.~\ref{fig:best ME} assesses how well the model sustains performance across different environments without fine-tuning. The results confirm that the proposed TL model generalizes effectively, reinforcing its robustness and practical applicability in dynamic industrial conditions.

\subsection{Limitations' Discussion and Future Work}
\label{lim}
Although the proposed TL-driven VLC localization framework achieves substantial gains in accuracy, energy, and data efficiency, several practical limitations and open questions remain that warrant explicit discussion and future study.

(i) \textbf{Adaptability under extreme environmental conditions:} The optical VLC channel is sensitive to extreme lighting conditions (direct sunlight, complete darkness, or rapid illumination transients) and to very dense multi-obstacle layouts that can reduce LOS reliability and increase positioning error \cite{gupta2024illuminating, sejan2023comprehensive}. These edge cases are not fully covered by our current experiments and represent opportunities for future investigation, particularly through hybrid sensing, optical-filtering or adaptive model strategies. Despite these challenges, the proposed framework holds potential for real-world applications, including smart logistics, industrial digital twins, and other factory automation scenarios where precise indoor localization is critical.

(ii) \textbf{Synchronization and deployment realism:} The measurement campaign used a VNA as a common frequency/time reference for TX and RX, which reduces synchronization-related variance relative to unsynchronized hardware. In field deployments with separate, unsynchronized TX and RX hardware, timing/frequency mismatches and small systematic offsets may increase measurement variance and bias feature values, potentially degrading localization accuracy. Practical mitigations include short calibration scans at a few known reference locations, insertion of pilot/reference tones for alignment, and per-scan normalization or offset correction of RSSI traces. We leave a systematic experimental quantification of these deployment effects to future work.

(iii) \textbf{Multi-sensor fusion:} Prior studies have shown that combining complementary sensors such as barometers, 3-D maps, geomagnetic sensors, and inertial measurements, can substantially enhance localization robustness and enable seamless indoor–outdoor operation \cite{li2024review,kuo2023intelligent,wang2023seamless}. These findings motivate exploring hybrid VLC-based approaches that integrate such modalities for improved adaptability and non-line-of-sight resilience. While our proposed work focuses solely on VLC-based TL, extending the framework toward multi-sensor fusion remains an open research direction. Several complementary fusion architectures could be investigated to further improve robustness and generalization: (a) \textit{feature-level fusion}, where synchronized barometric, geomagnetic, or inertial measurement unit (IMU) features can be combined with VLC features prior to fine-tuning; (b) \textit{probabilistic constraint fusion}, in which VLC model outputs can be integrated with auxiliary sensor priors using filtering or factor-graph optimization and (c) \textit{late or opportunistic fusion}, where estimates from different modalities are adaptively weighted based on signal quality or model confidence \cite{kaess2012isam2,li2024revieww}.

(iv) \textbf{Scalability and computational cost:} Scaling our proposed method to much larger VLC infrastructures or to more complex neural network architectures introduces synchronization, hierarchical-modeling, and data-management challenges that require systematic study and remain open for future investigation. In addition, fine-tuning directly on edge devices reduces centralized retraining but can impose CPU/GPU, memory, and energy burdens on edge hardware. Evaluating lightweight TL variants and measuring practical limits on representative edge platforms are left as important directions for future work.

(v) \textbf{Energy measurement:} For controlled and repeatable experiments we initially constrained the energy term to \textit{training-stage} consumption only; however, this excludes inference-stage and network communication energy which can be significant in deployed IoT scenarios. Training-stage energy was selected to ensure consistent comparisons across models and noise conditions, as inference energy can vary widely with deployment hardware and optimizations such as quantization, batching, or accelerators. Extending the analysis toward full life-cycle energy accounting—including training, inference, and communication phases—remains an open direction for future refinement \cite{mavromatis2024computing, chou2024energy}. 
\section{Conclusion}
\label{sec:conclusion}
In conclusion, this study presents a robust and efficient TL framework for VLC-based indoor localization in dynamic industrial environments. By developing a unified ME metric that encompasses performance accuracy, energy efficiency, data efficiency, and transfer gain, the proposed approach significantly improves localization performance. Specifically, it achieves up to 67.5\% ME and reduces localization error by up to 47\% compared to conventional models. Furthermore, it maintains high accuracy using only 30\% of the original VLC localization dataset collected from the production environment, highlighting its adaptability and scalability. Thus, the proposed framework holds significant potential for industrial automation and smart logistics. Future work will explore extending the framework's applicability through multi-sensor data fusion, deployment on edge devices for low-latency inference, and adaptation to increasingly complex and dynamic industrial layouts to enable real-time system operation.

\section*{Acknowledgment}

\noindent The authors gratefully acknowledge the financial supports of the EU Horizon 2020 program towards the 6G BRAINS project H2020-ICT 101017226.

\bibliographystyle{IEEEtran}
\bibliography{references}


\vspace{10pt}
\begin{IEEEbiography}[{\includegraphics[width=1in,height=1.25in, clip,keepaspectratio]{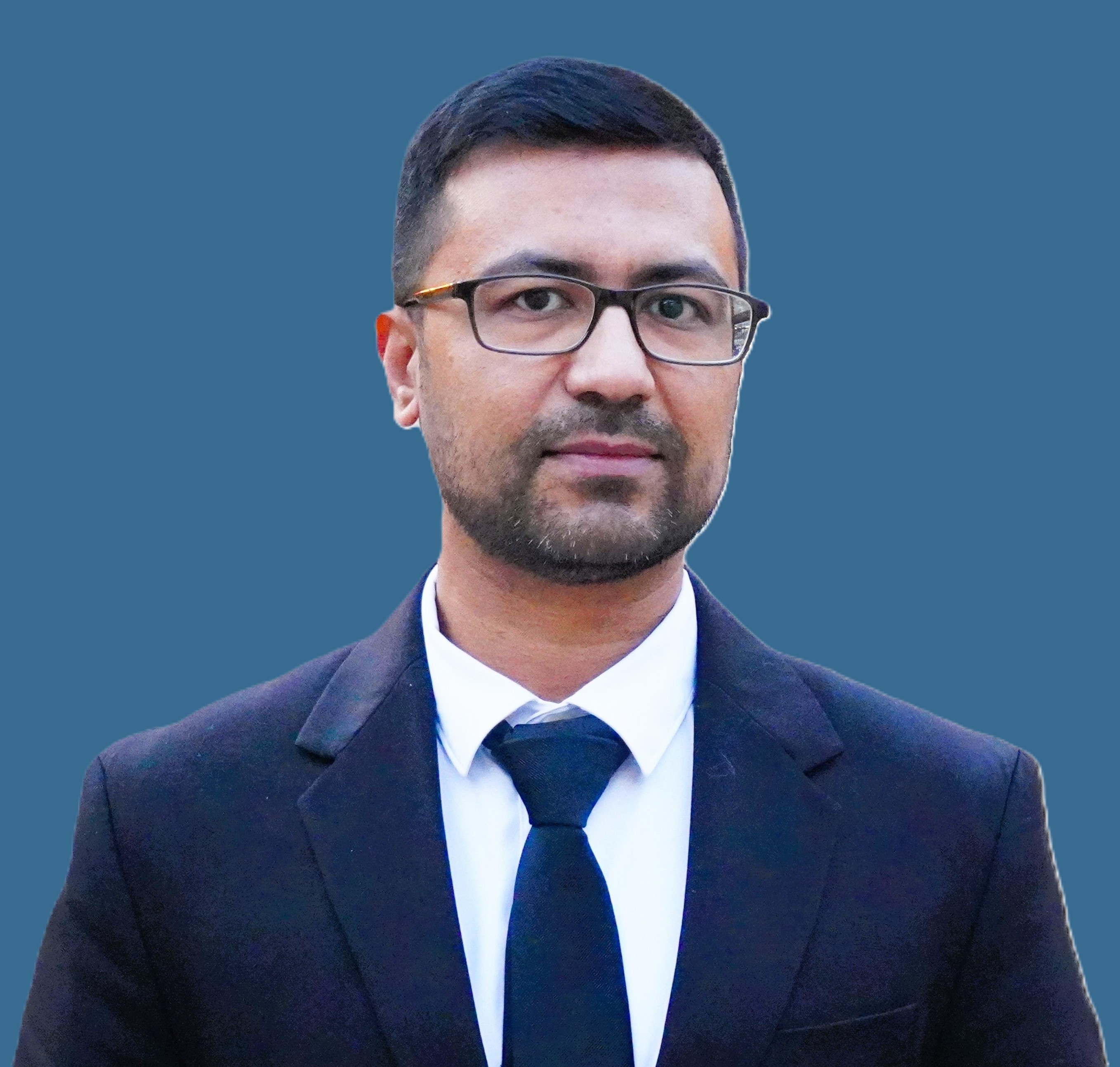}}]{Masood Jan} is a Ph.D. candidate in Machine Learning at Sorbonne University, Paris, France, and is affiliated with the LISITE Laboratory at ISEP, Paris. He was ranked first in the EDITE 2023 doctoral campaign. He received his Master's degree in Telecommunications and Networks from the Conservatoire National des Arts et Métiers (CNAM), Paris, France, in 2023, and the Eng. degree in Electronics Engineering from the University of Engineering \& Technology (UET), Peshawar, Pakistan, in 2008. Prior to his doctoral studies, he gained over 12 years of experience in the wireless communications industry, working with multinational companies on AI-driven analytics, automation, and network optimization for LTE/5G systems. His current research interests include machine learning in wireless communication and efficient AI models for 6G/OWC. He is actively engaged in academic dissemination and has delivered numerous talks at conferences and seminars.

\end{IEEEbiography}

\begin{IEEEbiography}[{\includegraphics[width=1in,height=1.25in, clip,keepaspectratio]{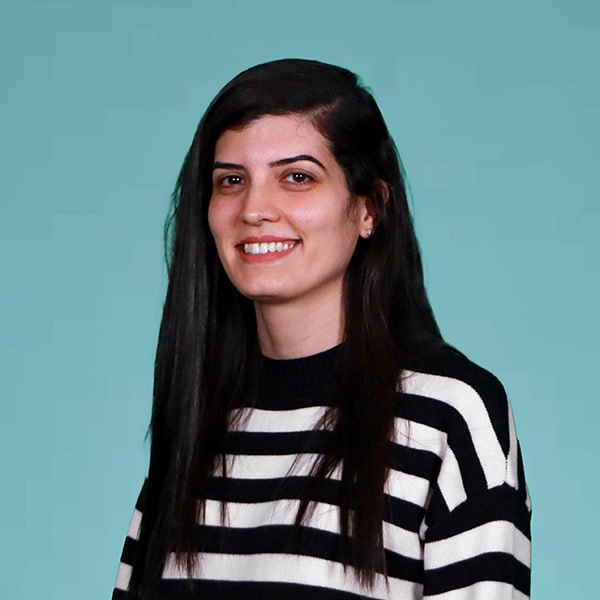}}]{Wafa Njima} (Member, IEEE) received the Eng. degree from the Institut National des Sciences Appliquées et de Technologies de Tunis in 2015. She received the Ph.D. degree in the field of radio-communications in 2019, from the Conservatoire National des Arts
et Métiers in Paris in collaboration with the Ecole Supérieure des Communications de Tunis. She joined the ETIS laboratory at ENSEA Cergy as a temporary assistant and researcher also as a postdoctoral researcher. She is currently an associate
professor at ISEP, Paris. Her publications span several research areas, and her research interests are related to several topics including signal processing, wireless communications, indoor localization, IoT and machine learning for communications. She is IEEE member and she served as TPC member and reviewer for many leading international conferences and journals.
\end{IEEEbiography}

\begin{IEEEbiography}[{\includegraphics[width=1in,height=1.25in, clip,keepaspectratio]{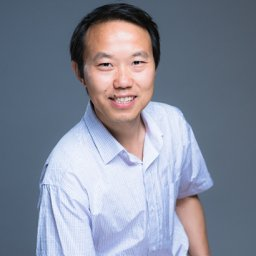}}]{Xun Zhang} (Senior Member, IEEE) is a Professor at ISEP (www.isep.fr), Paris, France, and conducts permanent research at the LISV Laboratory within the University of Saclay-Versailles. As a Senior Member of the IEEE, he serves as an Associate Editor for the IEEE Transactions on Broadcasting and is a member of the BTS Awards Committee. He earned his Ph.D. degree in Instrumentation and Microelectronics (IM) Science from the University of Nancy in October 2009. Between 2009 and 2011, Dr. Zhang undertook a Postdoctoral position at the SCEE Laboratory in CentraleSupélec, France, where he focused on applying auto-reconfigurable management methods and control strategies in FPGAs for cognitive radio systems. In 2011, Dr. Zhang joined the LISITE Laboratory at ISEP as an Associate Professor. His research primarily revolves around PHY layer optimization for 5G cellular networks and Visible Light Communication (VLC) systems. He is also pioneering the development of VLC-based indoor high-accuracy positioning and tracking algorithms.
\end{IEEEbiography}

\begin{IEEEbiography}[{\includegraphics[width=1in,height=1.25in, clip,keepaspectratio]{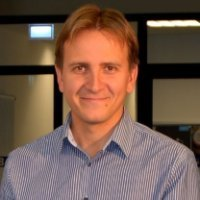}}]{Alexander Artemenko} is a research engineer within the Corporate Sector Research of Robert Bosch GmbH in Stuttgart, Germany. Prior to joining BOSCH in 2015, he was a postdoctoral researcher at the Ilmenau University of Technology (TU Ilmenau), Germany from 2013 to 2015. He received his PhD in engineering from the same university in 2013. In 2012, Dr. Artemenko was a post-doctoral researcher and lecturer at UCLA (Univ. of California Los Angeles) in the group of Professor Mario Gerla.
\end{IEEEbiography}
\vfill

\end{document}